\documentclass{article}

\usepackage{iclr2027_conference,times}
\usepackage{etoolbox}

\iclrfinalcopy

\makeatletter
\patchcmd{\@maketitle}
  {\lhead{Published as a conference paper at ICLR 2027}}
  {\lhead{Preprint. Under review.}}
  {}{}
\makeatother

\usepackage{amsmath,amsfonts,bm}

\def\eqref#1{equation~\ref{#1}}

\def\1{\bm{1}}

\DeclareMathAlphabet{\mathsfit}{\encodingdefault}{\sfdefault}{m}{sl}
\SetMathAlphabet{\mathsfit}{bold}{\encodingdefault}{\sfdefault}{bx}{n}

\usepackage[T1]{fontenc}
\usepackage[utf8]{inputenc}
\usepackage{latexsym}
\usepackage{amsmath}
\usepackage{microtype}
\usepackage{inconsolata}
\usepackage{graphicx}
\usepackage{booktabs}
\usepackage{multirow}
\usepackage{rotating}
\usepackage{colortbl}
\usepackage{hyperref}
\usepackage{url}
\usepackage{wrapfig}
\usepackage{float}
\usepackage{capt-of}
\usepackage{makecell}

\newcommand{\olmoe}{OLMoE}
\newcommand{\qwen}{Qwen1.5}
\newcommand{\deepseek}{DSV2}

\newcommand{\tolmoe}{OLMoE}
\newcommand{\tqwen}{Qwen1.5}
\newcommand{\tdeepseek}{DSV2}
\newcommand{\hlcell}[1]{\cellcolor{gray!15}#1}

\title{RA-MoE: Routing-Aligned Fine-Tuning for Multilingual Adaptation of Mixture-of-Experts Models}

\author{
\begin{tabular}{@{}l@{}}
{\normalfont\bfseries
Guanzhi Deng\textsuperscript{\rm 1},
Kuan Wu\textsuperscript{\rm 1},
Haibo Wang\textsuperscript{\rm 2},
Shing Yin Wong\textsuperscript{\rm 1},
Sichun Luo\textsuperscript{\rm 3},
}\\
{\normalfont\bfseries
Zengyan Liu\textsuperscript{\rm 1},
Fang Xu\textsuperscript{\rm 4},
Zhuoran Wang\textsuperscript{\rm 1},
Linqi Song\textsuperscript{\rm 1}\thanks{Corresponding author.}
}\\[6pt]
{\normalfont
\textsuperscript{\rm 1}City University of Hong Kong
\quad
\textsuperscript{\rm 2}Carnegie Mellon University
}\\
{\normalfont
\textsuperscript{\rm 3}The University of Hong Kong
\quad
\textsuperscript{\rm 4}Shenzhen Polytechnic University
}\\
{\normalfont
\texttt{guanzdeng2-c@my.cityu.edu.hk}
\quad
\texttt{linqi.song@cityu.edu.hk}
}
\end{tabular}
}

\iclrfinalcopy

\makeatletter

\patchcmd{\@maketitle}
  {\lhead{Published as a conference paper at ICLR 2027}}
  {\lhead{Preprint. Under review.}}
  {}{}

\patchcmd{\@maketitle}
  {\@title\par}
  {\@title\par\vskip 0.15in}
  {}{}

\makeatother

\begin{document}
\maketitle
\begin{abstract}
Mixture-of-Experts (MoE) models enable efficient LLM scaling, yet adapting
them to non-English downstream tasks remains challenging. Standard
multilingual fine-tuning largely ignores their heterogeneous routing
structure. Across multiple MoE models and tasks, we find strong
cross-lingual routing alignment in middle layers, with routing divergence
associated with target-language performance gaps. Motivated by
this observation, we propose \textbf{RA-MoE} (Routing-Aligned MoE
Fine-Tuning), a three-stage framework for multilingual MoE adaptation.
RA-MoE categorizes parallel examples into four correctness groups
(\textit{cc}/\textit{ci}/\textit{ic}/\textit{ii}) and identifies
task-relevant experts in middle layers. It then selectively aligns
target-language routing on \textit{ci} examples toward successful English
routing patterns, jointly matching the total routing mass assigned to
task experts and its relative allocation among them. Experiments across
three MoE models, three downstream tasks, and six target languages show
that RA-MoE consistently outperforms standard SFT and strong routing-aware
baselines. Further analyses confirm the intended routing changes and
reveal that middle-layer task routing is largely shared and transferable
across languages, providing mechanistic evidence for the cross-language
transferability of task-specific routing.
\end{abstract}

\section{Introduction}

\begin{wrapfigure}[18]{r}{0.42\textwidth}
    \vspace{-1.5em}
    \centering
    \includegraphics[width=\linewidth]{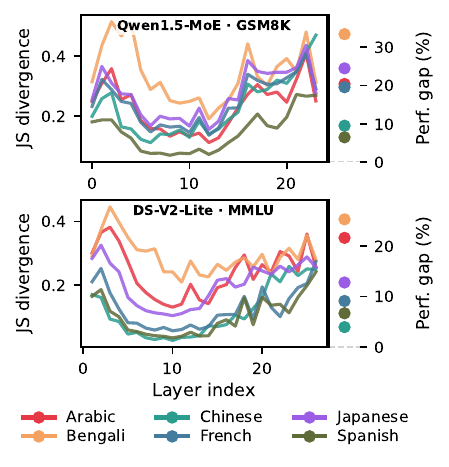}
    \vspace{-2.75em}
    \caption{Layer-wise cross-lingual routing divergence. Right-side
    markers show performance gaps from English.}
    \label{fig:ushape}
\end{wrapfigure}
Mixture-of-Experts (MoE) has emerged as a dominant paradigm for scaling
Large Language Models (LLMs) without proportionally increasing per-token
computation~\citep{shazeer2017outrageously, jiang2024mixtral,
liu2024deepseek}. Despite their impressive English-centric capabilities,
adapting MoE models to non-English downstream tasks remains
challenging. The prevailing approach is supervised fine-tuning (SFT) on
target-language task data~\citep{chen2024breaking, chai2025xcot,
zhang2026lingualift}, yet these methods are largely inherited from dense
LLMs and treat the model as a monolithic learner, without explicitly
exploiting the heterogeneous routing structure that MoEs develop during
pretraining.

A key clue lies in the routing dynamics of pretrained MoEs. Prior work
has revealed a characteristic U-shaped layer-wise pattern in cross-lingual
routing divergence: early and late layers tend to be more language-specific,
whereas middle layers exhibit substantially stronger cross-lingual
alignment~\citep{bandarkar2026multilingualroutingmixtureofexperts,
chen2026understandingmultilingualismmixtureofexpertsllms,
zheng2026unveilinglanguageroutingisolation}. We find that this structural
pattern also persists in task-specific settings.

As shown in Figure~\ref{fig:ushape}, the Jensen--Shannon divergence
between English and non-English routing distributions is consistently
higher in early and late layers and substantially reduced in the middle
layers across representative models and downstream tasks. Moreover,
languages with larger middle-layer routing divergence tend to exhibit
larger task performance gaps relative to English. These observations
suggest that middle layers constitute a particularly promising region for
cross-lingual routing intervention: they exhibit relatively language-shared
routing structure, while residual divergence in this region is associated
with degraded target-language task performance.

This motivates a natural question: \textit{can successful routing behavior
in English be used as an explicit training signal for adapting MoE models
to non-English tasks?} We propose \textbf{RA-MoE}
(\textbf{R}outing-\textbf{A}ligned \textbf{MoE} Fine-Tuning), a
three-stage framework. First, we run inference on parallel task examples
$(x_\text{en}, x_\text{tgt})$ and categorize them as
\textit{cc}/\textit{ci}/\textit{ic}/\textit{ii} according to correctness
in English and the target language. Second, we identify task-relevant
experts in the middle layers by contrasting routing behavior on successful
English task examples with that on general English data. Third, during
target-language SFT, we selectively align the routing of \textit{ci}
examples toward their successful English counterparts, matching both the
total routing mass assigned to task experts and its relative allocation
among them.

Experiments across three MoE models, three downstream tasks, and six
target languages show that RA-MoE consistently outperforms standard SFT
and strong routing-aware baselines. Further analyses confirm that these
gains are accompanied by the intended routing changes and reveal that
task-specific middle-layer routing is largely shared and transferable
across languages: successful non-English routing can also provide useful
supervision, while task-expert structures identified for one language can
be reused for others.

Our contributions are threefold:
(1) We validate the characteristic U-shaped cross-lingual routing
divergence pattern in task-specific settings and show that middle-layer
divergence is associated with target-language performance gaps.
(2) We propose \textbf{RA-MoE}, which combines correctness-based sample
selection, middle-layer task-expert identification, and selective routing
alignment that jointly matches task-expert routing mass and allocation.
(3) Across multiple MoE models, tasks, and languages, we demonstrate
consistent improvements over standard SFT and strong routing-aware
baselines, and further show that the underlying task-specific routing
structure is largely shared and transferable across languages.

\section{Related Work}

\noindent\textbf{Multilingual downstream task fine-tuning.}
The standard approach translates English task datasets into target
languages and applies SFT
\citep{chen2024breaking, chai2025xcot}, sometimes
augmented with cross-lingual representation alignment objectives
\citep{liu2025middle}. These methods improve non-English performance but
are primarily designed for dense LLMs, offering no mechanism to exploit
or align the sparse routing structure of MoEs.

\noindent\textbf{MoE-based multilingual language expansion.}
A distinct body of work uses MoE architectures to extend multilingual
capabilities of dense LLMs, including upcycling dense models into MoEs
with language-specific experts \citep{zhou2025moe}, layer-wise expert
allocation based on cross-lingual similarity \citep{zhang2025less}, and
neuron-level expert allocation
\citep{li2026neuronmoeneuronguidedmixtureofexpertsefficient}. These
methods use MoE to augment dense models with multilingual capacity,
whereas we focus on improving downstream multilingual performance of
models that are already MoE-based.

\noindent\textbf{Multilingual routing analysis in MoEs.}
Recent studies establish a U-shaped layer-wise pattern in
cross-lingual routing divergence, where middle layers exhibit stronger
cross-lingual alignment and their routing divergence correlates with
language performance
\citep{bandarkar2026multilingualroutingmixtureofexperts,
chen2026understandingmultilingualismmixtureofexpertsllms}. Both works
propose inference-time routing-guided steering but do not update model
parameters. \citet{zheng2026unveilinglanguageroutingisolation} further
reveal Language Routing Isolation between high- and low-resource
languages, and propose RISE to selectively fine-tune language-specific
expert subnetworks.

\noindent\textbf{Routing-aware MoE fine-tuning.}
RoMA \citep{li2025routingmanifoldalignmentimproves} shows that existing
MoE routers can be suboptimal on downstream tasks and improves
generalization by aligning each sample's routing to that of successful
neighbors. DR-LoRA \citep{deng2026drloradynamicranklora} addresses
capacity mismatch by dynamically allocating LoRA ranks based on routing
frequency and gradient signals.

While these lines of work collectively highlight the importance of
routing structure in multilingual MoE models, none explicitly uses
successful cross-lingual routing patterns as selective supervision for
non-English downstream adaptation. RA-MoE bridges this gap by identifying
task-relevant experts in the middle layers and selectively aligning
target-language routing toward successful English patterns on examples
exhibiting cross-lingual correctness asymmetry.

\section{Method}
\label{sec:method}

\begin{figure}[t]
    \centering
    \includegraphics[
        width=\linewidth,
        trim=0cm 4cm 0cm 3.2cm,
        clip
    ]{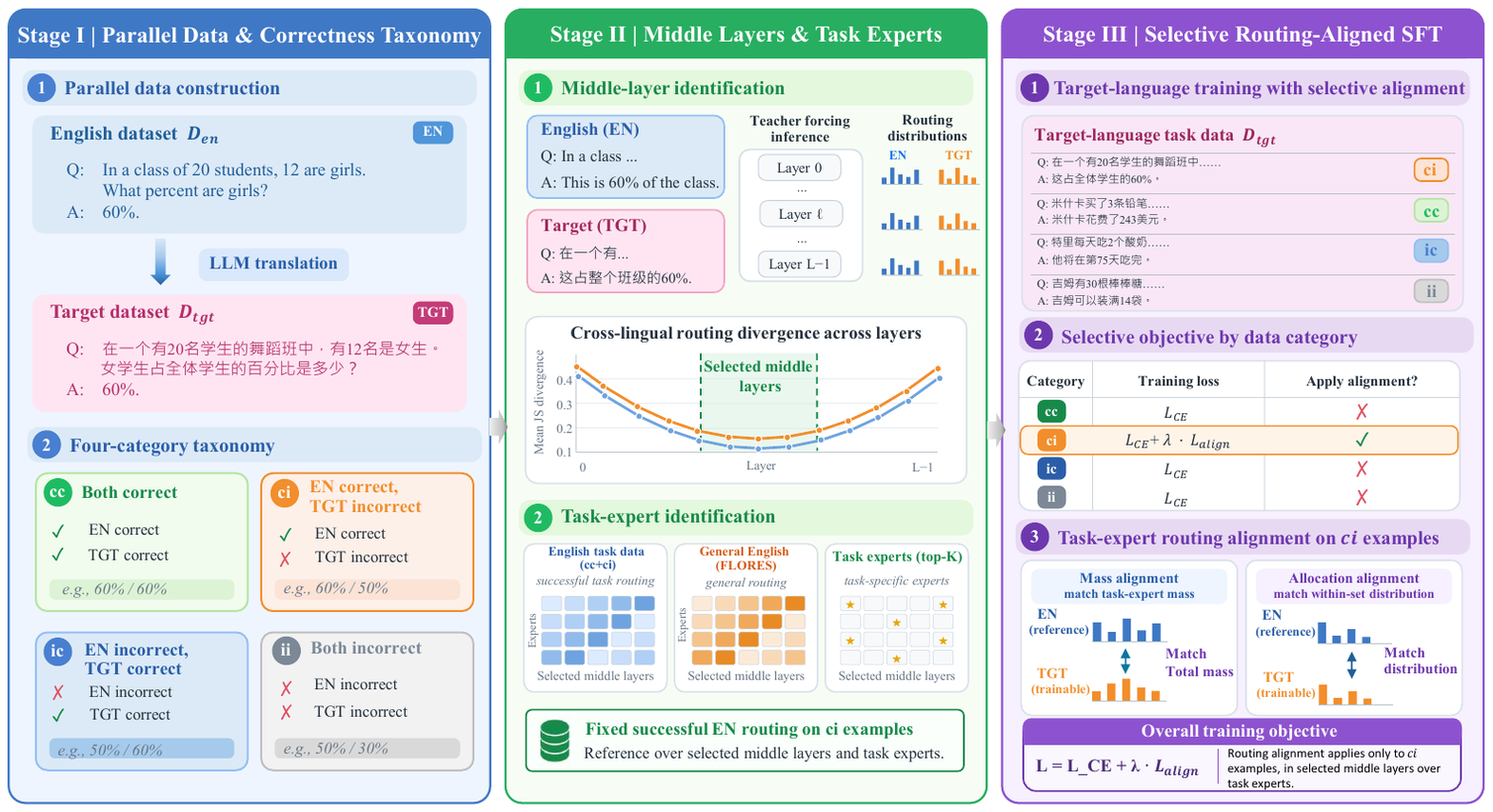}
    \vspace{-1em}
    \caption{Overview of the RA-MoE framework.}
    \label{fig:overview}
\end{figure}

\subsection{Overview}

RA-MoE builds on the task-specific cross-lingual routing pattern
established in Figure~\ref{fig:ushape}, where routing is substantially
more aligned across languages in the middle MoE layers, making these
layers a natural target for cross-lingual routing adaptation. As
illustrated in Figure~\ref{fig:overview}, RA-MoE proceeds in three stages.
\textbf{Stage 1} constructs parallel English--target-language task data
and partitions examples into four categories according to answer
correctness in the two languages.
\textbf{Stage 2} profiles routing behavior to identify the middle layers
with stronger cross-lingual alignment and the task-relevant experts
within them.
\textbf{Stage 3} performs target-language supervised fine-tuning while
selectively aligning the routing of \textit{ci} examples toward their
successful English counterparts, accounting for both the total routing
mass assigned to task experts and its normalized allocation within the
task-expert set.

\subsection{Stage 1: Parallel Data Construction and Categorization}

\noindent\textbf{Parallel data construction.}
Let
$\mathcal{D}_\text{en}
= \{(x_\text{en}^{(i)}, y_\text{en}^{(i)})\}_{i=1}^{N}$
denote an English downstream task dataset.
We construct its target-language counterpart
$\mathcal{D}_\text{tgt}
= \{(x_\text{tgt}^{(i)}, y_\text{tgt}^{(i)})\}_{i=1}^{N}$,
yielding parallel examples
$\{(x_\text{en}^{(i)},x_\text{tgt}^{(i)})\}_{i=1}^{N}$.

\noindent\textbf{Four-way correctness categorization.}
We run inference with the MoE model $\mathcal{M}$ on both
$x_\text{en}^{(i)}$ and $x_\text{tgt}^{(i)}$ and evaluate the resulting
responses $\hat{y}_\text{en}^{(i)}$ and
$\hat{y}_\text{tgt}^{(i)}$ against their gold answers.
Each parallel example is assigned to one of four categories:
\textbf{\textit{cc}} (both correct),
\textbf{\textit{ci}} (English correct, target incorrect),
\textbf{\textit{ic}} (English incorrect, target correct), or
\textbf{\textit{ii}} (both incorrect).
Of particular interest are \textit{ci} examples, which exhibit a clear
cross-lingual correctness asymmetry: the model successfully solves the
same underlying task in English but fails in the target language.
We use these examples as the targets for routing alignment in
Stage~3.

\subsection{Stage 2: Middle Layer and Task Expert Identification}

\noindent\textbf{Routing profiling.}
We profile routing on the model generations obtained in Stage~1.
For each parallel example, we run teacher-forcing inference on
$(x_\text{en}^{(i)},\hat{y}_\text{en}^{(i)})$ and
$(x_\text{tgt}^{(i)},\hat{y}_\text{tgt}^{(i)})$ and record the
post-softmax router distribution at each MoE layer.
For layer $\ell$, let
$\mathbf{p}_{t}^{(\text{lang},\ell,i)} \in [0,1]^E$
denote the routing distribution over $E$ experts for token $t$.
We average over generated token positions to obtain

\begin{equation}
    \mathbf{q}^{(\text{lang},\ell,i)}
    =
    \frac{1}{|G_\text{lang}^{(i)}|}
    \sum_{t \in G_\text{lang}^{(i)}}
    \mathbf{p}_{t}^{(\text{lang},\ell,i)},
\end{equation}

where $G_\text{lang}^{(i)}$ denotes the generated token positions for
example $i$ in language $\text{lang}$.

\noindent\textbf{Middle layer identification.}
We compute the mean cross-lingual routing divergence at each layer:

\begin{equation}
    \overline{\mathrm{Div}}^{(\ell)}
    =
    \frac{1}{N}\sum_{i=1}^{N}
    D_\text{JS}\!\left(
        \mathbf{q}^{(\text{en},\ell,i)}
        \,\Big\|\,
        \mathbf{q}^{(\text{tgt},\ell,i)}
    \right),
\end{equation}

where $D_\text{JS}$ denotes the entropy-normalized Jensen--Shannon
divergence.
Let $\tau$ denote the median of
$\{\overline{\mathrm{Div}}^{(\ell)}\}_{\ell}$.
We define $\mathcal{L}_\text{mid}=\{m,\ldots,n\}$ as the longest
contiguous range where
$\overline{\mathrm{Div}}^{(\ell)}\leq\tau$ for every
$\ell\in\mathcal{L}_\text{mid}$.
This adaptive criterion selects the low-divergence region
illustrated in Figure~\ref{fig:ushape}.

\noindent\textbf{Task expert identification.}
Within each $\ell\in\mathcal{L}_\text{mid}$, we identify experts preferentially activated when the model successfully solves the
downstream task in English.
We contrast correctly answered English task examples
$\mathcal{D}_\text{en}^\text{task}$, corresponding to the
\textit{cc} and \textit{ci} subsets, with a general English corpus
$\mathcal{D}_\text{en}^\text{gen}$ (FLORES-200;
\citealt{costa2022no}).
For each source, the mean routing weight of expert $e$ is

\begin{equation}
    \bar{f}_e^{(\text{src},\ell)}
    =
    \frac{1}{|\mathcal{D}_\text{src}|}
    \sum_{i\in\mathcal{D}_\text{src}}
    q_e^{(\text{src},\ell,i)},
\end{equation}

and its task-specificity score is

\begin{equation}
    \Delta_e^{(\ell)}
    =
    \bar{f}_e^{(\text{task},\ell)}
    -
    \bar{f}_e^{(\text{gen},\ell)}.
\end{equation}

We select the top-$K$ experts with
$\Delta_e^{(\ell)}>0$ as the task-expert set
$\mathcal{E}_\ell^\text{task}$.
In our experiments, we use $K=8$ and analyze sensitivity to $K$ in
Section~\ref{sec:ablation}.

For each \textit{ci} example, we additionally store its English routing
distributions
$\{\mathbf{q}^{(\text{en},\ell,i)}\}_{\ell\in\mathcal{L}_\text{mid}}$
as fixed reference signals for Stage~3.

\subsection{Stage 3: Routing-Aligned SFT}

\noindent\textbf{Routing alignment.}
For each \textit{ci} example $i$ and middle layer
$\ell\in\mathcal{L}_\text{mid}$, we align the target-language routing
with the corresponding successful English routing over the identified
task experts. Specifically, we obtain the target-language routing
distribution from the current SFT forward pass by averaging router
probabilities over response tokens, while keeping the English routing
distribution stored in Stage~2 fixed.

Let the total routing mass assigned to the task-expert set be

\begin{equation}
    m^{(\text{lang},\ell,i)}
    =
    \sum_{e\in\mathcal{E}_\ell^\text{task}}
    q_e^{(\text{lang},\ell,i)}.
\end{equation}

We construct a $(|\mathcal{E}_\ell^\text{task}|+1)$-dimensional routing
distribution that preserves the probability assigned to each task expert
while aggregating all remaining experts into a residual component:

\begin{equation}
    \mathbf{r}^{(\text{lang},\ell,i)}
    =
    \left[
        \left(
        q_e^{(\text{lang},\ell,i)}
        \right)_{e\in\mathcal{E}_\ell^\text{task}},
        \;
        1-m^{(\text{lang},\ell,i)}
    \right].
\end{equation}

The routing alignment loss averages the divergence across \textit{ci}
examples and selected middle layers:

\begin{equation}
    \mathcal{L}_\text{align}
    =
    \mathbb{E}_{i\in\mathcal{B}_\text{ci}}
    \mathbb{E}_{\ell\in\mathcal{L}_\text{mid}}
    \left[
    D_\text{KL}\!\left(
        \mathbf{r}^{(\text{en},\ell,i)}
        \,\Big\|\,
        \mathbf{r}^{(\text{tgt},\ell,i)}
    \right)
    \right],
\end{equation}

where $\mathcal{B}_\text{ci}$ denotes the \textit{ci} examples in the
current minibatch. The explicit minibatch estimator and implementation
details are provided in Appendix~\ref{app:impl}.

To characterize this objective, define the normalized task-expert
allocation
$\tilde{q}_e^{(\text{lang},\ell,i)}
=
q_e^{(\text{lang},\ell,i)}/m^{(\text{lang},\ell,i)}$
and the binary mass distribution
$\mathbf{b}^{(\text{lang},\ell,i)}
=
[m^{(\text{lang},\ell,i)},1-m^{(\text{lang},\ell,i)}]$.
The KL divergence decomposes as

\begin{equation}
D_\text{KL}\!\left(
\mathbf{r}^{(\text{en},\ell,i)} \|
\mathbf{r}^{(\text{tgt},\ell,i)}
\right)
=
D_\text{KL}\!\left(
\mathbf{b}^{(\text{en},\ell,i)} \|
\mathbf{b}^{(\text{tgt},\ell,i)}
\right)
+
m^{(\text{en},\ell,i)}
D_\text{KL}\!\left(
\tilde{\mathbf{q}}^{(\text{en},\ell,i)} \|
\tilde{\mathbf{q}}^{(\text{tgt},\ell,i)}
\right).
\end{equation}

The first term aligns the total routing mass assigned to task experts,
while the second aligns the normalized allocation within the task-expert
set, weighted by the English task-expert mass. A detailed derivation is
provided in Appendix~\ref{app:alignment_decomposition}.

\noindent\textbf{Combined objective.}
The final training objective is

\begin{equation}
    \mathcal{L}
    =
    \mathcal{L}_\text{CE}
    +
    \lambda\,\mathcal{L}_\text{align},
\end{equation}

where $\lambda$ controls the strength of routing alignment. Only
\textit{ci} examples contribute to $\mathcal{L}_\text{align}$, while
all training examples contribute to $\mathcal{L}_\text{CE}$.

\section{Experiments}

\subsection{Experimental Setup}

\noindent\textbf{Models.}
We evaluate RA-MoE on three MoE models of varying architectures and scales:
\textbf{OLMoE-1B-7B-0924-SFT}
(\citealt{muennighoff2025olmoe}, hereafter \olmoe), \textbf{Qwen1.5-MoE-A2.7B}
(\citealt{qwen_moe}, \qwen), and
\textbf{DeepSeek-V2-Lite-Chat}
(\citealt{deepseekai2024deepseekv2strongeconomicalefficient}, \deepseek).

\noindent\textbf{Tasks and languages.}
We evaluate on three downstream tasks:
\textbf{Mathematical Reasoning} (GSM8K;
\citealt{cobbe2021training}),
\textbf{Logical Reasoning} (LogicBench;
\citealt{parmar2024logicbench}), and
\textbf{Knowledge Understanding} (MMLU;
\citealt{hendrycks2021measuringmassivemultitasklanguage}).
For each task, we construct parallel English--target-language data in six
target languages: \textbf{Arabic} (ar), \textbf{Bengali} (bn),
\textbf{Chinese} (zh), \textbf{French} (fr), \textbf{Japanese} (ja),
and \textbf{Spanish} (es), covering a range of language families and
resource levels. Data construction and translation details are provided in
Appendix~\ref{app:data}.

\noindent\textbf{Baselines.}
We compare RA-MoE against four baselines:
\textbf{Zero-shot}, which evaluates the pretrained MoE directly on
target-language inputs without task-specific fine-tuning;
\textbf{Routing Steering}~\citep{bandarkar2026multilingualroutingmixtureofexperts},
an inference-time intervention that promotes English-aligned expert
activation in middle layers without updating model parameters;
\textbf{SFT}, which performs standard supervised fine-tuning on
target-language task data; and
\textbf{RISE}~\citep{zheng2026unveilinglanguageroutingisolation},
which selectively fine-tunes language-specific expert subnetworks identified
through routing analysis, without an explicit cross-lingual alignment signal. Implementation details are provided in
Appendix~\ref{app:impl}.

\subsection{Main Results}

\begin{table}[t]
\caption{Main results (\%) across three tasks, three models, and six target languages.
RS denotes Routing Steering \citep{bandarkar2026multilingualroutingmixtureofexperts}.
The \textbf{Avg.} column averages over all 18 task-language combinations.
Best result per column is \textbf{bolded}; second best is \underline{underlined}.}
\label{tab:main}
\centering
\renewcommand{\arraystretch}{1.15}
\setlength{\tabcolsep}{2pt}
\footnotesize
\resizebox{\linewidth}{!}{
\begin{tabular}{ll cccccc cccccc cccccc c}
\toprule
& & \multicolumn{6}{c}{\textbf{GSM8K}}
& \multicolumn{6}{c}{\textbf{LogicBench}}
& \multicolumn{6}{c}{\textbf{MMLU}} & \\
\cmidrule(lr){3-8} \cmidrule(lr){9-14} \cmidrule(lr){15-20}
\textbf{Model} & \textbf{Method}
  & \textbf{ar} & \textbf{bn} & \textbf{zh} & \textbf{fr} & \textbf{ja} & \textbf{es}
  & \textbf{ar} & \textbf{bn} & \textbf{zh} & \textbf{fr} & \textbf{ja} & \textbf{es}
  & \textbf{ar} & \textbf{bn} & \textbf{zh} & \textbf{fr} & \textbf{ja} & \textbf{es}
  & \textbf{Avg.} \\
\midrule

\multirow{5}{*}{\tolmoe}
& 0-shot
& 3.8\,{\tiny±0.1}
& 1.9\,{\tiny±0.2}
& 17.7\,{\tiny±0.4}
& 20.7\,{\tiny±0.4}
& 7.2\,{\tiny±0.3}
& 23.0\,{\tiny±0.2}
& 33.7\,{\tiny±0.5}
& 15.2\,{\tiny±0.4}
& 43.5\,{\tiny±1.2}
& 41.2\,{\tiny±0.6}
& 41.3\,{\tiny±0.7}
& 46.2\,{\tiny±0.5}
& 28.3\,{\tiny±0.1}
& 25.5\,{\tiny±0.1}
& 31.7\,{\tiny±0.1}
& 37.6\,{\tiny±0.1}
& 29.6\,{\tiny±0.1}
& 36.9\,{\tiny±0.1}
& 26.9\,{\tiny±0.3} \\

& RS
& 5.1\,{\tiny±0.1}
& 2.0\,{\tiny±0.1}
& 18.6\,{\tiny±0.3}
& 21.3\,{\tiny±0.3}
& 6.8\,{\tiny±0.3}
& 23.6\,{\tiny±0.2}
& 34.3\,{\tiny±0.4}
& 15.5\,{\tiny±0.2}
& 43.2\,{\tiny±0.4}
& 41.6\,{\tiny±0.3}
& 41.3\,{\tiny±0.5}
& 46.5\,{\tiny±0.4}
& 27.5\,{\tiny±0.1}
& 25.0\,{\tiny±0.1}
& 31.3\,{\tiny±0.1}
& 37.1\,{\tiny±0.1}
& 29.5\,{\tiny±0.1}
& 37.3\,{\tiny±0.1}
& 27.1\,{\tiny±0.2} \\

& SFT
& 9.4\,{\tiny±0.2}
& \textbf{2.2\,{\tiny±0.1}}
& 21.3\,{\tiny±0.3}
& \underline{25.7\,{\tiny±0.3}}
& 13.8\,{\tiny±0.3}
& \underline{28.3\,{\tiny±0.2}}
& 34.9\,{\tiny±0.6}
& 15.7\,{\tiny±0.4}
& 44.5\,{\tiny±0.7}
& 42.8\,{\tiny±0.7}
& 43.5\,{\tiny±1.1}
& 48.8\,{\tiny±0.9}
& 28.4\,{\tiny±0.1}
& 25.6\,{\tiny±0.1}
& 32.1\,{\tiny±0.1}
& 38.0\,{\tiny±0.1}
& 29.7\,{\tiny±0.1}
& 37.8\,{\tiny±0.1}
& 29.0\,{\tiny±0.4} \\

& RISE
& \underline{9.7\,{\tiny±0.2}}
& 2.0\,{\tiny±0.1}
& \underline{21.7\,{\tiny±0.3}}
& \underline{25.7\,{\tiny±0.3}}
& \underline{14.2\,{\tiny±0.3}}
& 28.1\,{\tiny±0.2}
& \underline{35.7\,{\tiny±0.8}}
& \underline{16.3\,{\tiny±0.4}}
& \underline{45.8\,{\tiny±0.7}}
& \underline{43.6\,{\tiny±1.0}}
& \underline{45.3\,{\tiny±0.8}}
& \underline{50.5\,{\tiny±0.7}}
& \underline{29.1\,{\tiny±0.1}}
& \underline{25.8\,{\tiny±0.1}}
& \underline{32.9\,{\tiny±0.1}}
& \underline{38.2\,{\tiny±0.1}}
& \underline{29.9\,{\tiny±0.1}}
& \underline{38.1\,{\tiny±0.1}}
& \underline{29.6\,{\tiny±0.4}} \\

\rowcolor{gray!15}
& \textbf{RA-MoE}
& \textbf{10.8\,{\tiny±0.2}}
& \textbf{2.2\,{\tiny±0.2}}
& \textbf{22.7\,{\tiny±0.3}}
& \textbf{26.9\,{\tiny±0.3}}
& \textbf{15.3\,{\tiny±0.5}}
& \textbf{29.8\,{\tiny±0.5}}
& \textbf{37.5\,{\tiny±0.4}}
& \textbf{17.2\,{\tiny±0.6}}
& \textbf{47.8\,{\tiny±0.8}}
& \textbf{46.3\,{\tiny±0.5}}
& \textbf{47.2\,{\tiny±0.6}}
& \textbf{51.7\,{\tiny±0.7}}
& \textbf{30.6\,{\tiny±0.1}}
& \textbf{27.7\,{\tiny±0.1}}
& \textbf{33.6\,{\tiny±0.1}}
& \textbf{39.3\,{\tiny±0.1}}
& \textbf{31.5\,{\tiny±0.1}}
& \textbf{39.2\,{\tiny±0.1}}
& \textbf{31.0\,{\tiny±0.3}} \\

\midrule

\multirow{5}{*}{\tqwen}
& 0-shot
& 29.3\,{\tiny±0.5}
& 3.2\,{\tiny±0.2}
& 28.8\,{\tiny±0.3}
& 41.8\,{\tiny±0.5}
& 9.4\,{\tiny±0.6}
& 44.4\,{\tiny±0.4}
& 59.6\,{\tiny±0.8}
& 64.4\,{\tiny±0.3}
& 63.1\,{\tiny±0.7}
& 53.9\,{\tiny±1.5}
& 60.6\,{\tiny±0.7}
& 66.8\,{\tiny±0.4}
& 35.8\,{\tiny±0.1}
& 24.3\,{\tiny±0.1}
& 46.7\,{\tiny±0.1}
& 45.2\,{\tiny±0.1}
& 37.9\,{\tiny±0.1}
& 45.4\,{\tiny±0.1}
& 42.3\,{\tiny±0.4} \\

& RS
& 29.5\,{\tiny±0.5}
& 4.1\,{\tiny±0.2}
& 29.4\,{\tiny±0.3}
& 41.3\,{\tiny±0.5}
& 9.9\,{\tiny±0.5}
& 45.1\,{\tiny±0.4}
& 60.2\,{\tiny±0.6}
& 63.5\,{\tiny±0.5}
& 63.4\,{\tiny±0.3}
& 53.5\,{\tiny±0.8}
& 61.2\,{\tiny±0.4}
& 66.2\,{\tiny±0.7}
& 36.4\,{\tiny±0.1}
& 24.9\,{\tiny±0.1}
& 46.0\,{\tiny±0.1}
& 45.8\,{\tiny±0.1}
& 38.3\,{\tiny±0.1}
& 45.1\,{\tiny±0.1}
& 42.4\,{\tiny±0.4} \\

& SFT
& 48.8\,{\tiny±0.4}
& 29.5\,{\tiny±0.6}
& 56.7\,{\tiny±0.3}
& 57.2\,{\tiny±0.6}
& 42.3\,{\tiny±0.4}
& 58.5\,{\tiny±0.6}
& 61.1\,{\tiny±0.9}
& 65.5\,{\tiny±0.5}
& 65.2\,{\tiny±0.9}
& 55.8\,{\tiny±1.1}
& 62.6\,{\tiny±1.3}
& 68.5\,{\tiny±0.6}
& 40.0\,{\tiny±0.1}
& 28.3\,{\tiny±0.1}
& 52.8\,{\tiny±0.1}
& 50.6\,{\tiny±0.1}
& 42.7\,{\tiny±0.1}
& 51.2\,{\tiny±0.1}
& 52.1\,{\tiny±0.5} \\

& RISE
& \underline{50.1\,{\tiny±0.3}}
& \underline{30.6\,{\tiny±0.6}}
& \underline{57.5\,{\tiny±0.3}}
& \underline{57.7\,{\tiny±0.6}}
& \underline{43.4\,{\tiny±0.4}}
& \underline{59.0\,{\tiny±0.5}}
& \underline{62.0\,{\tiny±0.7}}
& \underline{66.1\,{\tiny±0.4}}
& \underline{66.2\,{\tiny±0.6}}
& \underline{56.8\,{\tiny±1.2}}
& \underline{63.7\,{\tiny±0.8}}
& \underline{69.2\,{\tiny±0.8}}
& \underline{40.3\,{\tiny±0.1}}
& \underline{28.6\,{\tiny±0.1}}
& \underline{53.2\,{\tiny±0.1}}
& \underline{51.0\,{\tiny±0.1}}
& \textbf{43.1\,{\tiny±0.1}}
& \underline{52.3\,{\tiny±0.1}}
& \underline{52.8\,{\tiny±0.4}} \\

\rowcolor{gray!15}
& \textbf{RA-MoE}
& \textbf{52.0\,{\tiny±0.5}}
& \textbf{32.9\,{\tiny±0.5}}
& \textbf{59.8\,{\tiny±0.3}}
& \textbf{60.2\,{\tiny±0.7}}
& \textbf{46.6\,{\tiny±0.7}}
& \textbf{61.3\,{\tiny±0.5}}
& \textbf{63.4\,{\tiny±0.6}}
& \textbf{67.2\,{\tiny±1.1}}
& \textbf{67.5\,{\tiny±0.5}}
& \textbf{58.4\,{\tiny±0.8}}
& \textbf{65.3\,{\tiny±0.3}}
& \textbf{70.4\,{\tiny±0.8}}
& \textbf{42.1\,{\tiny±0.1}}
& \textbf{30.3\,{\tiny±0.1}}
& \textbf{55.2\,{\tiny±0.1}}
& \textbf{52.5\,{\tiny±0.1}}
& \textbf{43.1\,{\tiny±0.1}}
& \textbf{53.9\,{\tiny±0.1}}
& \textbf{54.6\,{\tiny±0.4}} \\

\midrule

\multirow{5}{*}{\tdeepseek}
& 0-shot
& 26.0\,{\tiny±0.6}
& 10.2\,{\tiny±0.5}
& 63.7\,{\tiny±0.3}
& 58.7\,{\tiny±0.5}
& 43.7\,{\tiny±0.6}
& 61.4\,{\tiny±0.5}
& 54.6\,{\tiny±0.7}
& 47.7\,{\tiny±1.1}
& 57.6\,{\tiny±1.0}
& 54.8\,{\tiny±0.6}
& 53.7\,{\tiny±0.8}
& 53.3\,{\tiny±0.9}
& 32.1\,{\tiny±0.1}
& 28.8\,{\tiny±0.1}
& 49.9\,{\tiny±0.1}
& 44.7\,{\tiny±0.1}
& 40.8\,{\tiny±0.1}
& 47.4\,{\tiny±0.1}
& 46.1\,{\tiny±0.5} \\

& RS
& 26.4\,{\tiny±0.5}
& 11.6\,{\tiny±0.4}
& 63.1\,{\tiny±0.3}
& \underline{59.3\,{\tiny±0.5}}
& 44.2\,{\tiny±0.5}
& 61.7\,{\tiny±0.4}
& 55.0\,{\tiny±0.6}
& 48.2\,{\tiny±0.5}
& 57.1\,{\tiny±0.5}
& 56.1\,{\tiny±0.7}
& 53.9\,{\tiny±0.3}
& 54.3\,{\tiny±0.2}
& 32.5\,{\tiny±0.1}
& 28.9\,{\tiny±0.1}
& 49.3\,{\tiny±0.1}
& 44.4\,{\tiny±0.1}
& 41.1\,{\tiny±0.1}
& 47.9\,{\tiny±0.1}
& 46.4\,{\tiny±0.3} \\

& SFT
& 37.6\,{\tiny±0.4}
& 19.4\,{\tiny±0.4}
& 64.2\,{\tiny±0.2}
& 58.6\,{\tiny±0.4}
& 50.0\,{\tiny±0.4}
& 63.4\,{\tiny±0.1}
& 55.5\,{\tiny±0.9}
& 49.1\,{\tiny±0.6}
& 58.6\,{\tiny±0.7}
& 58.6\,{\tiny±1.0}
& 55.9\,{\tiny±0.9}
& 58.8\,{\tiny±1.0}
& 32.3\,{\tiny±0.1}
& 29.2\,{\tiny±0.1}
& 50.0\,{\tiny±0.1}
& 45.0\,{\tiny±0.1}
& 42.2\,{\tiny±0.1}
& 48.2\,{\tiny±0.1}
& 48.7\,{\tiny±0.4} \\

& RISE
& \underline{37.9\,{\tiny±0.3}}
& \underline{19.9\,{\tiny±0.4}}
& \underline{64.4\,{\tiny±0.2}}
& 58.9\,{\tiny±0.4}
& \underline{50.3\,{\tiny±0.4}}
& \underline{63.6\,{\tiny±0.2}}
& \underline{57.0\,{\tiny±0.8}}
& \underline{50.5\,{\tiny±0.7}}
& \underline{60.9\,{\tiny±1.1}}
& \underline{60.2\,{\tiny±0.7}}
& \underline{57.2\,{\tiny±0.6}}
& \underline{61.5\,{\tiny±0.9}}
& \underline{32.9\,{\tiny±0.1}}
& \underline{29.3\,{\tiny±0.1}}
& \underline{50.1\,{\tiny±0.1}}
& \underline{45.2\,{\tiny±0.1}}
& \underline{42.5\,{\tiny±0.1}}
& \underline{48.3\,{\tiny±0.1}}
& \underline{49.5\,{\tiny±0.4}} \\

\rowcolor{gray!15}
& \textbf{RA-MoE}
& \textbf{39.4\,{\tiny±0.6}}
& \textbf{20.8\,{\tiny±0.3}}
& \textbf{65.6\,{\tiny±0.4}}
& \textbf{61.4\,{\tiny±0.6}}
& \textbf{53.1\,{\tiny±0.5}}
& \textbf{64.6\,{\tiny±0.2}}
& \textbf{58.9\,{\tiny±1.3}}
& \textbf{52.8\,{\tiny±0.8}}
& \textbf{63.3\,{\tiny±0.7}}
& \textbf{62.6\,{\tiny±0.8}}
& \textbf{59.3\,{\tiny±0.5}}
& \textbf{63.5\,{\tiny±0.9}}
& \textbf{35.0\,{\tiny±0.1}}
& \textbf{30.5\,{\tiny±0.1}}
& \textbf{52.2\,{\tiny±0.1}}
& \textbf{48.0\,{\tiny±0.1}}
& \textbf{43.7\,{\tiny±0.1}}
& \textbf{49.9\,{\tiny±0.1}}
& \textbf{51.4\,{\tiny±0.5}} \\

\bottomrule
\end{tabular}}
\end{table}

Table~\ref{tab:main} presents the main results across three MoE models,
three tasks, and six target languages. We report mean accuracy and standard deviation over five
evaluation runs using different random seeds.

Several observations emerge. First, RA-MoE consistently improves over
standard SFT, outperforming it in 53 of the 54 model--task--language
settings and tying in the remaining one, with an average gain of 2.4
percentage points. The improvement holds across all three tasks:
RA-MoE gains 2.1 points on GSM8K, 3.1 on LogicBench, and 1.9 on MMLU
on average. In particular, it improves over SFT in all 18 LogicBench and
all 18 MMLU settings, while achieving 17 wins and one tie on GSM8K.

Second, RA-MoE substantially outperforms Routing Steering across all 54
settings, with an average gain of 7.0 points. While Routing Steering
directly modifies expert routing at inference time, its improvements over
the zero-shot model are limited in many cases. In contrast, incorporating
routing supervision during fine-tuning yields substantially stronger
adaptation across reasoning, logical, and knowledge-intensive tasks.

Third, RA-MoE consistently improves over RISE, outperforming it in
53 of 54 settings and tying once, with an average gain of 1.7 points.
The advantage is observed across all three tasks: RA-MoE outperforms RISE
in every GSM8K and LogicBench setting, and in 17 of 18 MMLU settings.
These results indicate that explicitly aligning target-language routing
toward successful English task-routing patterns provides gains
beyond selectively adapting language-specific expert subnetworks.

\subsection{Ablation and Sensitivity Analysis}
\label{sec:ablation}

\begin{figure}[H]
\begin{minipage}[t]{0.52\textwidth}
\vspace{0pt}
\vspace{-\abovecaptionskip}
\captionof{table}{Ablation study on \qwen{} (GSM8K).}
\label{tab:ablation}

\centering
\renewcommand{\arraystretch}{1.12}
\setlength{\tabcolsep}{3.2pt}
\scriptsize
\begin{tabular}{lccc}
\toprule
\textbf{Method} & \textbf{bn} & \textbf{ar} & \textbf{es} \\
\midrule
w/o $\mathcal{L}_\text{align}$
& 29.5\,{\tiny±0.6}
& 48.8\,{\tiny±0.4}
& 58.5\,{\tiny±0.6} \\

w/o mass alignment
& 31.8\,{\tiny±0.6}
& 50.7\,{\tiny±0.3}
& 60.4\,{\tiny±0.5} \\

w/o allocation alignment
& \underline{32.2\,{\tiny±0.5}}
& \underline{51.3\,{\tiny±0.4}}
& \underline{60.6\,{\tiny±0.6}} \\

w/o task experts
& 31.5\,{\tiny±0.5}
& 50.5\,{\tiny±0.3}
& 59.4\,{\tiny±0.7} \\

w/o \textit{ci} filtering
& 31.4\,{\tiny±0.4}
& 50.2\,{\tiny±0.4}
& 59.3\,{\tiny±0.5} \\

w/o middle layers
& 30.2\,{\tiny±0.6}
& 49.4\,{\tiny±0.6}
& 58.5\,{\tiny±0.3} \\

\midrule
\rowcolor{gray!15}
\textbf{RA-MoE}
& \textbf{32.9\,{\tiny±0.5}}
& \textbf{52.0\,{\tiny±0.5}}
& \textbf{61.3\,{\tiny±0.5}} \\
\bottomrule
\end{tabular}
\vspace{-1em}
\end{minipage}
\hfill
\begin{minipage}[t]{0.46\textwidth}
\vspace{-0.25em}
\centering
\includegraphics[
    width=\linewidth,
    trim=0 8 0 8,
    clip
]{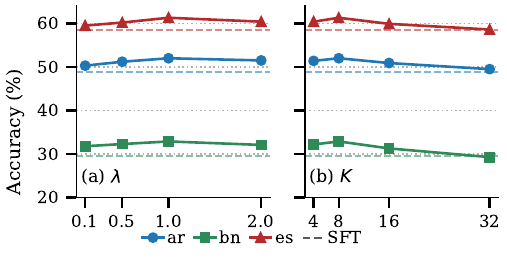}
\vspace{-2em}
\captionof{figure}{Sensitivity analysis on \qwen{} (GSM8K).
Dashed lines denote SFT baselines.}
\vspace{-1em}
\label{fig:sensitivity}
\end{minipage}
\end{figure}

\noindent\textbf{Ablation.}
To assess the contribution of each component of RA-MoE, we conduct
ablations on \qwen{} with GSM8K in Bengali, Arabic, and Spanish.
Table~\ref{tab:ablation} compares the full method with six variants:
\textbf{(1) w/o $\mathcal{L}_\text{align}$}, removing routing alignment;
\textbf{(2) w/o mass alignment} and
\textbf{(3) w/o allocation alignment}, removing the mass and within-set
allocation components, respectively;
\textbf{(4) w/o task experts}, aligning all experts in the selected
middle layers;
\textbf{(5) w/o \textit{ci} filtering}, aligning all training examples;
and \textbf{(6) w/o middle layers}, aligning all MoE layers.

As shown in Table~\ref{tab:ablation}, all ablations degrade performance
relative to the full RA-MoE. Removing routing alignment causes the
largest overall drop. Among the two alignment components, retaining only
mass alignment consistently outperforms retaining only allocation
alignment, while combining both achieves the best results, indicating
their complementary roles. The remaining ablations further confirm the
importance of selectively aligning task experts, \textit{ci} examples,
and middle MoE layers.

\noindent\textbf{Sensitivity to $\lambda$ and $K$.}
Figure~\ref{fig:sensitivity} reports performance as a function of the
alignment weight $\lambda \in \{0.1, 0.5, 1.0, 2.0\}$ and the number of
task experts per layer $K \in \{4, 8, 16, 32\}$ on \qwen{} (GSM8K).
RA-MoE remains robust across a broad range of both hyperparameters.
Too small a $\lambda$ weakens the routing alignment signal, whereas an
overly large $\lambda$ can interfere with the cross-entropy objective.
Similarly, too small a $K$ may omit relevant experts, while too large a
$K$ can include less task-specific experts and dilute the alignment
target.

\vspace{-1em}
\noindent

\section{Further Analysis}
\label{sec:analysis}

We analyze when RA-MoE is most beneficial, whether its gains
are accompanied by the intended routing changes, and how broadly the
underlying task-routing structure transfers across languages.

\subsection{Higher \textit{ci} Proportions Correlate with Larger Gains}
\label{sec:analysis_ci}

\begin{figure}[H]
\begin{minipage}[t]{0.59\textwidth}
\vspace{0pt}
This trend is consistent with the selective design of RA-MoE:
\textit{ci} examples capture cases where the model succeeds in English
but fails on the same task in the target language, making the successful
English routing pattern an informative alignment reference. By contrast,
\textit{cc} examples offer less corrective signal, while \textit{ii}
examples lack a successful English reference. Thus, task--language pairs
with more \textit{ci} examples provide more actionable routing
supervision. OLMoE on Bengali is a notable outlier, showing little gain
despite a high \textit{ci} proportion, suggesting that this signal is
most informative when the base model already has sufficient
target-language capability.
\end{minipage}
\hfill
\begin{minipage}[t]{0.38\textwidth}
\vspace{-0.5em}
\centering
\includegraphics[width=\linewidth]{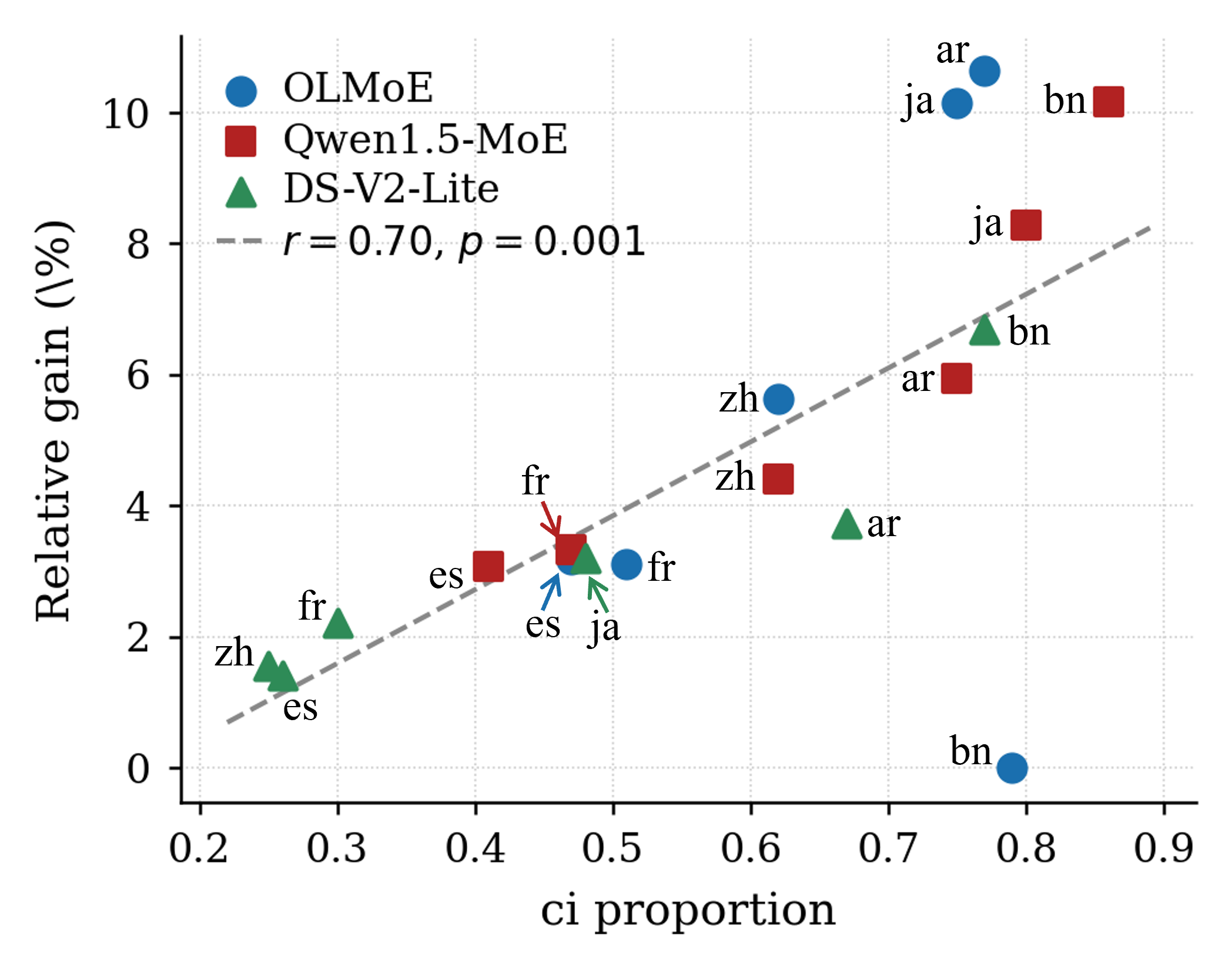}
\vspace{-2.5em}
\caption{Relative gain of RA-MoE over SFT vs.\ \textit{ci} proportion
on GSM8K.}
\label{fig:ci_proportion}
\end{minipage}
\vspace{-1.25em}
\end{figure}

We examine whether the proportion of \textit{ci} examples is associated
with the
benefit of RA-MoE over standard SFT. As shown in Figure~\ref{fig:ci_proportion}, the relative gain
over SFT is positively correlated with the \textit{ci} proportion across
languages and models ($r=0.70$, $p=0.001$).

\subsection{RA-MoE Induces the Intended Routing Changes}
\label{sec:routing_analysis}

Having identified when RA-MoE tends to yield larger gains, we next
examine whether these gains are accompanied by the intended routing
changes. Specifically, we analyze cross-lingual routing divergence before
and after fine-tuning and track task-expert activation throughout training.

\begin{figure}[H]
    \centering
    \begin{minipage}[t]{0.57\textwidth}
        \vspace{0pt}
        \centering
        \includegraphics[width=\linewidth]{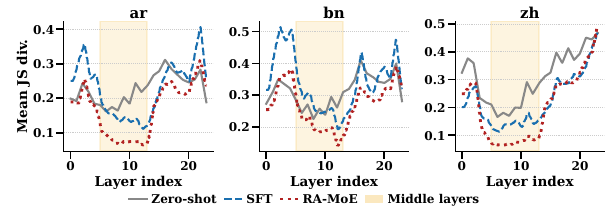}
        \vspace{-0.8em}

        {\footnotesize (a)}
    \end{minipage}
    \hfill
    \begin{minipage}[t]{0.40\textwidth}
        \vspace{0pt}
        \centering
        \includegraphics[width=\linewidth]{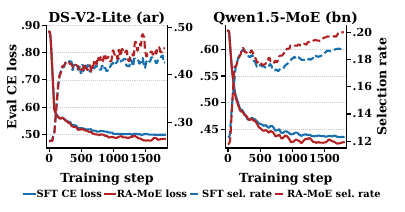}
        \vspace{-0.8em}

        {\footnotesize (b)}
    \end{minipage}
    \vspace{-0.5em}
    \caption{Routing-level analysis of RA-MoE.
    (a) Layer-wise cross-lingual routing divergence on Qwen1.5 (GSM8K);
    shading marks selected middle layers.
    (b) Eval CE loss and task-expert selection rate
    (fraction of routed top-$k$ slots assigned to task experts)
    under SFT and RA-MoE.}
    \label{fig:routing_analysis}
\end{figure}

\textbf{RA-MoE substantially reduces middle-layer routing divergence.}
As shown in Figure~\ref{fig:routing_analysis}(a), before fine-tuning,
all three target languages exhibit pronounced divergence from English,
including within the selected middle layers. Standard SFT reduces this
divergence only modestly, whereas RA-MoE consistently brings Arabic,
Bengali, and Chinese routing closer to the English reference. This shows
that the performance improvements of RA-MoE are accompanied by the
intended change in cross-lingual routing behavior.

\textbf{RA-MoE increases task-expert activation while lowering eval CE loss.}
Figure~\ref{fig:routing_analysis}(b) tracks these two signals throughout
training on two representative model--language pairs. Under both methods,
task-expert activation increases as training progresses, but RA-MoE
reaches a consistently higher selection rate. Meanwhile, its eval CE
loss becomes lower than that of SFT after the early training stage.
Together, these trends suggest that routing alignment encourages stronger
engagement of the task-relevant experts identified in Stage~2 while
improving target-language task adaptation.

\subsection{Task-Specific Routing Transfers Across Languages}
\label{sec:cross_lingual_transfer}

Beyond examining how RA-MoE changes routing during adaptation, we next
ask whether task-specific routing structure transfers across languages.
We study this question from two complementary perspectives: whether
successful routing supervision can come from non-English reference
languages, and whether the middle-layer task-expert structure identified
for one target language can be reused for others.

\vspace{0.3em}
\noindent
\begin{minipage}[t]{0.38\textwidth}
\vspace{0pt}
\vspace{-\abovecaptionskip}

\captionof{table}{Effect of reference language.}
\label{tab:reference_language}

\centering
\renewcommand{\arraystretch}{0.90}
\setlength{\tabcolsep}{2.5pt}
\scriptsize
\resizebox{\linewidth}{!}{%
\begin{tabular}{llccc}
\toprule
\textbf{Model}
& \textbf{Ref.}
& \textbf{Ref. Acc.}
& \textbf{Target Acc.}
& \textbf{Gain} \\
\midrule

\multirow{4}{*}{\tdeepseek{}}
& SFT & --   & 50.0\,{\tiny±0.4} & -- \\
& en  & 72.3 & 53.1\,{\tiny±0.5} & +3.1 \\
& zh  & 64.2 & 51.7\,{\tiny±0.4} & +1.7 \\
& es  & 63.4 & 51.2\,{\tiny±0.4} & +1.2 \\

\midrule

\multirow{4}{*}{\tqwen{}}
& SFT & --   & 42.3\,{\tiny±0.4} & -- \\
& en  & 88.5 & 46.6\,{\tiny±0.7} & +4.3 \\
& fr  & 57.2 & 43.8\,{\tiny±0.5} & +1.5 \\
& es  & 58.5 & 44.5\,{\tiny±0.4} & +2.2 \\

\bottomrule
\end{tabular}}
\end{minipage}
\hfill
\begin{minipage}[t]{0.59\textwidth}
\vspace{0pt}
\vspace{-\abovecaptionskip}

\captionof{table}{Cross-language transfer of task experts.}
\label{tab:cross_lang}

\centering
\renewcommand{\arraystretch}{1.08}
\setlength{\tabcolsep}{1.5pt}
\scriptsize
\resizebox{\linewidth}{!}{%
\begin{tabular}{ll cc cc cc c}
\toprule
& & \multicolumn{2}{c}{\textbf{GSM8K}}
& \multicolumn{2}{c}{\textbf{LogicBench}}
& \multicolumn{2}{c}{\textbf{MMLU}}
& \multirow{2}{*}{\makecell{\textbf{Avg. Gain}\\\textbf{over SFT}}} \\
\cmidrule(lr){3-4}
\cmidrule(lr){5-6}
\cmidrule(lr){7-8}
\textbf{Model} & \textbf{Method}
& \textbf{ar} & \textbf{bn}
& \textbf{ar} & \textbf{bn}
& \textbf{ar} & \textbf{bn}
& \\
\midrule

\multirow{2}{*}{\tolmoe}
& RA-MoE
& 10.8\,{\tiny±0.2}
& 2.2\,{\tiny±0.2}
& 37.5\,{\tiny±0.4}
& 17.2\,{\tiny±0.6}
& 30.6\,{\tiny±0.1}
& 27.7\,{\tiny±0.1}
& +1.6 \\

&
\hlcell{RA-MoE$^\dagger$}
& \hlcell{10.2\,{\tiny±0.2}}
& \hlcell{2.1\,{\tiny±0.1}}
& \hlcell{36.0\,{\tiny±0.5}}
& \hlcell{16.6\,{\tiny±0.4}}
& \hlcell{29.2\,{\tiny±0.2}}
& \hlcell{26.5\,{\tiny±0.1}}
& \hlcell{+0.7} \\

\midrule

\multirow{2}{*}{\tqwen}
& RA-MoE
& 52.0\,{\tiny±0.5}
& 32.9\,{\tiny±0.5}
& 63.4\,{\tiny±0.6}
& 67.2\,{\tiny±1.1}
& 42.1\,{\tiny±0.1}
& 30.3\,{\tiny±0.1}
& +2.5 \\

&
\hlcell{RA-MoE$^\dagger$}
& \hlcell{51.1\,{\tiny±0.5}}
& \hlcell{31.9\,{\tiny±0.4}}
& \hlcell{62.5\,{\tiny±0.6}}
& \hlcell{66.0\,{\tiny±0.9}}
& \hlcell{40.8\,{\tiny±0.1}}
& \hlcell{29.0\,{\tiny±0.1}}
& \hlcell{+1.4} \\

\midrule

\multirow{2}{*}{\tdeepseek}
& RA-MoE
& 39.4\,{\tiny±0.6}
& 20.8\,{\tiny±0.3}
& 58.9\,{\tiny±1.3}
& 52.8\,{\tiny±0.8}
& 35.0\,{\tiny±0.1}
& 30.5\,{\tiny±0.1}
& +2.4 \\

&
\hlcell{RA-MoE$^\dagger$}
& \hlcell{38.8\,{\tiny±0.5}}
& \hlcell{20.2\,{\tiny±0.2}}
& \hlcell{57.5\,{\tiny±0.8}}
& \hlcell{51.2\,{\tiny±0.7}}
& \hlcell{33.3\,{\tiny±0.1}}
& \hlcell{29.5\,{\tiny±0.1}}
& \hlcell{+1.2} \\

\bottomrule
\end{tabular}}
\end{minipage}

\noindent\textbf{Successful routing supervision transfers beyond English.}
We first ask whether English is uniquely required as the routing
reference. We consider two GSM8K settings, both targeting Japanese:
\deepseek{} with English, Chinese, or Spanish as the reference, and
\qwen{} with English, French, or Spanish as the reference. Each non-English
reference language is first adapted with standard SFT to ensure
sufficient task competence before its routing patterns are used for
alignment. In both settings, the non-English reference languages
substantially outperform Japanese on the task.

As shown in Table~\ref{tab:reference_language}, all non-English
references improve target-language performance over standard SFT,
showing that effective routing supervision is not specific to English.
A language with substantially stronger task competence can also serve as
a useful reference for a weaker target language. However, non-English
references yield smaller gains than English, which has the largest
performance advantage over the target language. This suggests that
routing-transfer effectiveness is associated with the reference--target
task-competence gap.

\noindent\textbf{Task experts can be reused across target languages.}
We next examine whether the task-routing structure identified for one
target language can be transferred to others. For each task, we directly
reuse the middle-layer range and task-expert sets identified under the
Chinese target-language configuration when fine-tuning Arabic or
Bengali. We denote this setting as RA-MoE$^\dagger$.

As shown in Table~\ref{tab:cross_lang}, RA-MoE$^\dagger$ preserves
positive average gains over SFT across all three models, although its
improvements are smaller than those of standard RA-MoE. Thus,
middle-layer task-expert structure can be reused across target languages
without re-running the identification stage.

\subsection{Mechanistic Evidence for Language-Shared Task Routing}
\label{sec:shared_task_routing}

The transfer results above show that routing supervision and task-expert
structure can both transfer across languages. We
next ask what routing structure underlies this transferability. To this
end, we compare task experts identified for GSM8K across English,
Chinese, and Spanish on \deepseek{}. For routing-pattern comparisons, we
use a shared set of parallel examples correctly solved in all three
reference languages but incorrectly solved in Japanese, controlling for
sample composition.

\begin{figure}[t]
    \centering
    \includegraphics[width=\textwidth]{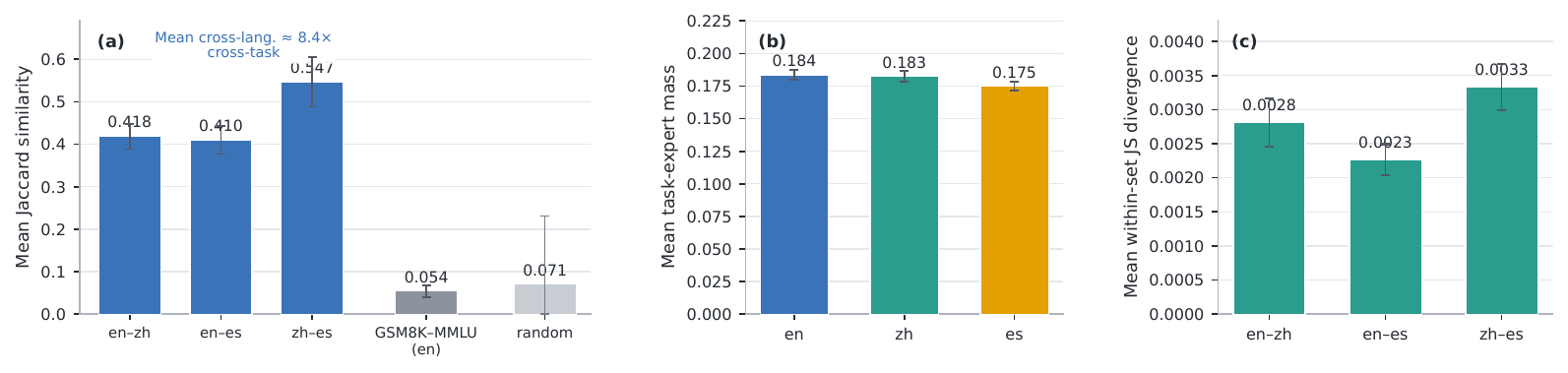}
    \vspace{-2em}
    \caption{Language-shared middle-layer task routing on
    \deepseek{} (GSM8K). (a) Task-expert overlap with cross-task and
    random controls. (b) Mean task-expert mass. (c) Within-set
    JS divergence.}
    \label{fig:task_expert_consistency}
\end{figure}

\noindent\textbf{Task-expert identity is substantially more stable across
languages than across tasks.}
For two task-expert sets $\mathcal{E}_A^{(\ell)}$ and
$\mathcal{E}_B^{(\ell)}$ at layer $\ell$, we measure their overlap using
Jaccard similarity:
$J^{(\ell)}(A,B)=
\frac{|\mathcal{E}_A^{(\ell)}\cap\mathcal{E}_B^{(\ell)}|}
{|\mathcal{E}_A^{(\ell)}\cup\mathcal{E}_B^{(\ell)}|}$. As shown in Figure~\ref{fig:task_expert_consistency}(a), the mean
cross-language Jaccard similarity on GSM8K is 0.458 across English,
Chinese, and Spanish, compared with only 0.054 between GSM8K and MMLU
when the language is fixed to English. The same-task cross-language
overlap is therefore about 8.4$\times$ higher than the cross-task
overlap. It also substantially exceeds the random-expert baseline
(0.071 for $E=64$ and $K=8$); even the smallest observed cross-language Jaccard (0.410) is matched by
the random baseline with an empirical probability of only
$3.57\times10^{-4}$. These results suggest that
middle-layer expert identity is considerably more task-dependent than
language-dependent.

\noindent\textbf{Task-expert usage is also similar across languages.}
Figure~\ref{fig:task_expert_consistency}(b) shows that English, Chinese,
and Spanish assign very similar mean routing mass to the shared GSM8K
task experts (0.184, 0.183, and 0.175, respectively), with mean
pairwise differences below 0.011. Figure~\ref{fig:task_expert_consistency}(c)
shows that their normalized routing distributions within this
shared set are highly similar, with pairwise JS divergence of only
0.0023--0.0033. Thus, successful execution of the same task across
languages is characterized not only by overlapping expert identity but
also by similar routing patterns over those experts.

Together, these results provide mechanistic evidence for a
language-shared view of middle-layer task routing: once the task is
fixed, successful execution across languages tends to recruit a common
set of task-relevant experts and use them in similar ways. Cross-lingual
performance gaps may therefore arise partly from how effectively each
language accesses this shared task-expert structure, rather than from
requiring a fundamentally different set of experts.

\subsection{Extension to Proxy-Based Sample Selection}
\label{sec:proxy_selection}

\noindent
\begin{minipage}[t]{0.55\textwidth}
\vspace{0pt}

Our main experiments focus on tasks for which training examples can be
categorized using objective answer correctness, allowing
\textit{cc}/\textit{ci}/\textit{ic}/\textit{ii} to be defined without
auxiliary quality proxies. We further examine whether RA-MoE can be
extended to settings where exact correctness labels are unavailable for
training-time sample selection.

\end{minipage}
\hfill
\begin{minipage}[t]{0.42\textwidth}
\vspace{0pt}
\vspace{-\abovecaptionskip}

\captionof{table}{Proxy-based extension on IFEval.}
\label{tab:proxy_ifeval}

\centering
\renewcommand{\arraystretch}{1.05}
\setlength{\tabcolsep}{10.0pt}
\scriptsize
\begin{tabular}{lccc}
\toprule
\textbf{Model}
& \textbf{SFT}
& \hlcell{\textbf{RA-MoE}}
& \textbf{Gain} \\
\midrule

\tolmoe{}
& 21.1\,{\tiny$\pm$0.4}
& \hlcell{\textbf{22.8\,{\tiny$\pm$0.4}}}
& +1.7 \\

\tqwen{}
& 24.8\,{\tiny$\pm$0.4}
& \hlcell{\textbf{27.0\,{\tiny$\pm$0.4}}}
& +2.2 \\

\tdeepseek{}
& 24.2\,{\tiny$\pm$0.3}
& \hlcell{\textbf{26.5\,{\tiny$\pm$0.4}}}
& +2.3 \\

\midrule

\textbf{Avg.}
& 23.4
& \hlcell{\textbf{25.4}}
& +2.1 \\

\bottomrule
\end{tabular}

\end{minipage}

\vspace{0.4em}

We consider IFEval, whose downstream performance can still be evaluated
objectively, while its open-ended instruction-following training data do
not provide per-response correctness labels. We therefore replace exact
\textit{ci} labels with a PPL-derived \textit{proxy-ci} criterion.
For each parallel pair, we compute perplexity over assistant-response
tokens for the English and target-language responses,
$\mathrm{PPL}_{\mathrm{en}}$ and $\mathrm{PPL}_{\mathrm{tgt}}$, and
define
$\Delta=\mathrm{PPL}_{\mathrm{tgt}}-\mathrm{PPL}_{\mathrm{en}}$.
An example is selected as \textit{proxy-ci} when
$\mathrm{PPL}_{\mathrm{en}}$ is below its corpus-level median and
$\Delta$ is above its corpus-level median. We use this criterion only
as a noisy sample-selection signal rather than a ground-truth
correctness label.

RA-MoE with proxy-based
selection improves over standard SFT in all 18 model--language
settings, with an average gain of 2.1 points as shown in Table~\ref{tab:proxy_ifeval}. This suggests
that exact correctness labels are not strictly required when an
informative relative-quality signal is available: \textit{ci} selection
serves to identify a reliable reference-stronger/target-weaker
asymmetry, for which response likelihood can provide an approximate
signal. A manual inspection of 600 \textit{proxy-ci} examples across
six target languages finds that 85.2\% match the intended pattern,
while 7.1\% are false positives caused by translation or formatting
artifacts and 7.7\% are ambiguous. Thus, although imperfect,
PPL-derived selection captures the intended cross-lingual competence
asymmetry in most cases and remains a useful, albeit noisy, alignment
signal.

\section{Conclusion}

We presented RA-MoE, a routing-aligned fine-tuning framework
for multilingual adaptation of MoE models. Motivated by the task-specific
U-shaped pattern of cross-lingual routing divergence, RA-MoE selectively
aligns target-language routing on \textit{ci} examples toward successful
English task-expert patterns in the middle layers, jointly matching
task-expert routing mass and within-set allocation. Experiments across
three MoE models, three downstream tasks, and six target languages show
consistent improvements over standard SFT and strong routing-aware
baselines. Further analyses confirm the intended routing changes, show
that routing supervision and task-expert structure can transfer across
languages, and demonstrate that informative proxy signals can extend
RA-MoE beyond settings with exact correctness labels. Overall, our
findings suggest that successful task execution across languages relies
on a largely shared middle-layer routing structure. Future work may
explore richer quality signals for open-ended tasks and adaptive selection
of reference languages.

\bibliographystyle{iclr2027_conference}
\bibliography{ref}

\appendix


\section{Data Construction}
\label{app:data}

\subsection{Core Training and Evaluation Datasets}
\label{app:core_data}

Our core experiments cover three downstream tasks with objectively
verifiable answers: mathematical reasoning, logical reasoning, and
knowledge understanding. For each task, we construct parallel
English--target-language training examples in Arabic (ar), Bengali (bn),
Chinese (zh), French (fr), Japanese (ja), and Spanish (es). Table~\ref{tab:datasets}
summarizes the training/evaluation pairings used in the main experiments.

\noindent\textbf{Mathematical reasoning.}
We sample 60{,}000 examples from MetaMathQA~\citep{yu2024metamath}, a
dataset of rewritten and augmented GSM8K and MATH problems. Models are
evaluated on the GSM8K test set~\citep{cobbe2021training} using
exact-match accuracy with an 8-shot prompt.

\noindent\textbf{Logical reasoning.}
We use LogicBench~\citep{parmar2024logicbench} for logical reasoning,
using LogicBench (Aug) for training and LogicBench (Eval) for evaluation.
The task admits objective answer labels, allowing exact correctness-based
categorization in Stage~1.

\noindent\textbf{Knowledge understanding.}
We evaluate on MMLU~\citep{hendrycks2021measuringmassivemultitasklanguage}
using 0-shot accuracy. For adaptation, we sample 60{,}000 examples from
the MMLU auxiliary/train split as the English training data, while the
MMLU test set is reserved for evaluation in the six target languages.

\begin{table}[ht]
\caption{Training and evaluation dataset pairings for the three core downstream tasks.}
\label{tab:datasets}
\centering
\small
\renewcommand{\arraystretch}{1.12}
\setlength{\tabcolsep}{3.5pt}
\resizebox{\columnwidth}{!}{%
\begin{tabular}{llllcc}
\toprule
\textbf{Task} & \textbf{Train set} & \textbf{Eval set}
& \textbf{Metric} & \textbf{Few-shot} & \textbf{Train size} \\
\midrule
Math Reasoning
& MetaMathQA
& GSM8K
& Exact-match acc.
& 8-shot
& 60{,}000 \\
Logical Reasoning
& LogicBench Aug-BQA
& LogicBench Eval-BQA
& Accuracy
& 0-shot
& 12{,}908 \\
Knowledge Understanding
& MMLU auxiliary train
& MMLU test
& Accuracy
& 0-shot
& 60{,}000 \\
\bottomrule
\end{tabular}}
\end{table}

\subsection{Translation Pipeline}
\label{app:translation}

\noindent\textbf{Translation model.}
We translate English training and evaluation data into the six target
languages using \texttt{google/gemini-2.5-flash-lite} accessed through
the OpenRouter API, where applicable.

\noindent\textbf{Dataset-specific handling.}
For MetaMathQA, numerical expressions, equations, and chain-of-thought
answer templates are preserved verbatim during translation to avoid
introducing arithmetic errors. For LogicBench, option labels, symbolic
entities, and task-specific constraint templates are kept unchanged to
preserve the underlying logical structure and answer space. For GSM8K evaluation, we translate the original English test split into
the six target languages using the same translation pipeline. For MMLU,
we sample the English adaptation data from the auxiliary/train split and
translate them into the six target languages through OpenRouter, while
evaluation uses the existing multilingual MMLU benchmark rather than
automatically translating the English test set.

\noindent\textbf{Translation quality.}
We use COMETKiwi~\citep{rei-etal-2022-cometkiwi}, a reference-free neural
quality-estimation model, to assess translated splits. Table~\ref{tab:comet_quality} reports translation-quality scores for
all datasets translated through OpenRouter in our experiments.

\begin{table}[t]
\caption{COMETkiwi translation-quality scores for datasets translated
through OpenRouter across six target languages. The \textbf{Role} column
indicates how each dataset is used in our experiments. Higher is better.}
\label{tab:comet_quality}

\centering
\small
\renewcommand{\arraystretch}{1.10}
\setlength{\tabcolsep}{3.5pt}

\begin{tabular}{llcccccc}
\toprule
\textbf{Dataset}
& \textbf{Role}
& \textbf{ar}
& \textbf{bn}
& \textbf{zh}
& \textbf{fr}
& \textbf{ja}
& \textbf{es} \\
\midrule

MetaMathQA
& GSM8K train
& 0.88 & 0.85 & 0.88 & 0.87 & 0.87 & 0.86 \\

GSM8K (test)
& GSM8K eval
& 0.83 & 0.88 & 0.85 & 0.87 & 0.87 & 0.86 \\

LogicBench (Aug \& Eval)
& LogicBench train/eval
& 0.87 & 0.84 & 0.87 & 0.86 & 0.85 & 0.85 \\

MMLU (auxiliary/train)
& MMLU train
& 0.88 & 0.87 & 0.85 & 0.87 & 0.86 & 0.88 \\

OLMoE-SFT-Mix
& IFEval train
& 0.84 & 0.87 & 0.86 & 0.87 & 0.87 & 0.86 \\

IFEval
& IFEval eval
& 0.89 & 0.85 & 0.88 & 0.87 & 0.85 & 0.86 \\

\bottomrule
\end{tabular}
\end{table}

\subsection{Exact Correctness Annotation for the Core Tasks}
\label{app:correctness}

The three core tasks use objective correctness criteria to define the
\textit{cc}/\textit{ci}/\textit{ic}/\textit{ii} categories. We apply
the same task-specific evaluation procedure to the English and
target-language inputs for each parallel example.

\noindent\textbf{Mathematical reasoning (MetaMathQA/GSM8K).}
We run greedy decoding on the English and target-language versions of
each training example and extract the final numerical answer using a
regex-based extractor supporting common answer formats (e.g.,
\texttt{\#\#\#\#} delimiters, \texttt{\textbackslash boxed\{\}}, and
language-specific answer phrases). A response is marked correct if its
extracted answer matches the gold answer after numeric normalization.

\noindent\textbf{Logical reasoning (LogicBench).}
For both Aug-BQA and Eval-BQA, the valid prediction space is
$\{\texttt{yes},\texttt{no}\}$. We first search the generated response,
case-insensitively, for occurrences matching
\texttt{Final answer: yes} or \texttt{Final answer: no}, allowing
arbitrary whitespace around the colon, and use the last occurrence if
multiple matches are found. Otherwise, we fall back to the last
standalone occurrence of \texttt{yes} or \texttt{no} in the response.
If neither label can be extracted, the prediction is treated as empty and
counted as incorrect. English gold labels are stripped and lowercased.
For translated Eval-BQA data, language-specific affirmative and negative
gold labels are deterministically mapped to the canonical
\texttt{yes}/\texttt{no} labels before comparison using a fixed mapping
for Arabic, Bengali, Chinese, French, Japanese, and Spanish. Model
responses are evaluated in this canonical label space.

\noindent\textbf{Knowledge understanding (MMLU).}
Training-time MMLU correctness is computed as a constrained
multiple-choice decision over
$\{\texttt{A},\texttt{B},\texttt{C},\texttt{D}\}$ rather than by
parsing a free-form generated response. After a language-specific answer
prefix, we perform a single-token, zero-temperature decoding step and
obtain the next-token log probability for each of the four option-letter
tokens. The option with the highest log probability is selected and
compared exactly with the gold answer. Integer gold labels, when present,
are mapped positionally as
$0\mapsto\texttt{A}$, $1\mapsto\texttt{B}$,
$2\mapsto\texttt{C}$, and $3\mapsto\texttt{D}$.
No normalization of spelled-out, language-specific, parenthesized, or
otherwise free-form answer surfaces is required because prediction is
restricted directly to the four option tokens.

\subsection{Proxy-Based Sample Selection for IFEval}
\label{app:proxy_ifeval}

Section~\ref{sec:proxy_selection} examines whether RA-MoE can be extended
to settings where exact correctness labels are unavailable for
training-time sample selection. We use the OLMoE SFT Mix as open-ended
instruction-following training data and evaluate on
IFEval~\citep{zhou2023instruction} using prompt-level loose accuracy.
Unlike the three core tasks, this extension uses a response-likelihood proxy to identify examples
for routing alignment.

\noindent\textbf{PPL-derived \textit{proxy-ci}.}
For each parallel training pair, we compute per-sample perplexity over
assistant-response tokens only, excluding user-prompt tokens. Let
$\mathrm{PPL}_{\mathrm{en}}$ and $\mathrm{PPL}_{\mathrm{tgt}}$ denote
the English and target-language perplexities, and define
\begin{equation}
    \Delta_{\mathrm{PPL}}
    =
    \mathrm{PPL}_{\mathrm{tgt}}
    -
    \mathrm{PPL}_{\mathrm{en}}.
\end{equation}
An example is selected as \textit{proxy-ci} when
$\mathrm{PPL}_{\mathrm{en}}$ is below its corpus-level median and
$\Delta_{\mathrm{PPL}}$ is above its corpus-level median. Samples with
undefined PPL (e.g., zero valid assistant tokens) or PPL above the 99th
percentile are excluded before computing the thresholds. This criterion
is intended to identify a reference-stronger/target-weaker asymmetry,
rather than literal correctness.

\noindent\textbf{IFEval translation and evaluation.}
Because the original IFEval verifier assumes English punctuation
conventions, we adapt sentence-boundary and comma-based constraints to
the six target languages. Sentence-ending punctuation is extended to
the corresponding native conventions, and comma detection recognizes
both Latin and script-specific commas. Translation-quality scores
for both the OLMoE SFT Mix training data and the translated IFEval
evaluation set are reported in Table~\ref{tab:comet_quality}.

\noindent\textbf{Full IFEval results.}
The main paper reports results averaged over the six target languages.
Table~\ref{tab:proxy_ifeval_full} provides the full 18
model--language comparisons underlying the finding that proxy-based
RA-MoE outperforms SFT in every setting.

\begin{table}[t]
\caption{Full IFEval results (prompt-level loose accuracy, \%) for the
proxy-based extension.}
\label{tab:proxy_ifeval_full}
\centering
\small
\renewcommand{\arraystretch}{1.05}
\setlength{\tabcolsep}{2.8pt}

\begin{tabular}{llccccccc}
\toprule
\textbf{Model} & \textbf{Method}
& \textbf{ar} & \textbf{bn} & \textbf{zh} & \textbf{fr}
& \textbf{ja} & \textbf{es} & \textbf{Avg.} \\
\midrule

\multirow{2}{*}{\tolmoe}
& SFT
& 18.1\,{\tiny±0.4} & 14.1\,{\tiny±0.4} & 17.8\,{\tiny±0.3}
& 27.8\,{\tiny±0.4} & 21.3\,{\tiny±0.3} & 27.5\,{\tiny±0.3}
& 21.1\,{\tiny±0.4} \\

& \hlcell{RA-MoE}
& \hlcell{19.8\,{\tiny±0.4}}
& \hlcell{16.0\,{\tiny±0.3}}
& \hlcell{19.3\,{\tiny±0.4}}
& \hlcell{29.6\,{\tiny±0.3}}
& \hlcell{23.0\,{\tiny±0.4}}
& \hlcell{29.3\,{\tiny±0.3}}
& \hlcell{22.8\,{\tiny±0.4}} \\

\midrule

\multirow{2}{*}{\tqwen}
& SFT
& 21.4\,{\tiny±0.5} & 20.5\,{\tiny±0.4} & 22.3\,{\tiny±0.5}
& 28.9\,{\tiny±0.3} & 25.8\,{\tiny±0.4} & 29.6\,{\tiny±0.4}
& 24.8\,{\tiny±0.4} \\

& \hlcell{RA-MoE}
& \hlcell{24.5\,{\tiny±0.3}}
& \hlcell{22.2\,{\tiny±0.4}}
& \hlcell{24.5\,{\tiny±0.4}}
& \hlcell{31.1\,{\tiny±0.4}}
& \hlcell{27.7\,{\tiny±0.4}}
& \hlcell{32.0\,{\tiny±0.4}}
& \hlcell{27.0\,{\tiny±0.4}} \\

\midrule

\multirow{2}{*}{\tdeepseek}
& SFT
& 18.2\,{\tiny±0.4} & 15.3\,{\tiny±0.3} & 24.6\,{\tiny±0.4}
& 30.5\,{\tiny±0.3} & 26.4\,{\tiny±0.3} & 30.0\,{\tiny±0.2}
& 24.2\,{\tiny±0.3} \\

& \hlcell{RA-MoE}
& \hlcell{20.6\,{\tiny±0.4}}
& \hlcell{17.7\,{\tiny±0.4}}
& \hlcell{26.9\,{\tiny±0.4}}
& \hlcell{32.7\,{\tiny±0.3}}
& \hlcell{28.6\,{\tiny±0.4}}
& \hlcell{32.3\,{\tiny±0.3}}
& \hlcell{26.5\,{\tiny±0.4}} \\

\bottomrule
\end{tabular}
\end{table}

\noindent\textbf{Human validation of \textit{proxy-ci}.}
We manually inspect 600 randomly sampled \textit{proxy-ci} examples,
with 100 examples from each target language. Annotators proficient in
both the target language and English assign each example to one of three
categories: (i) valid \textit{proxy-ci}, where the English response is
acceptable and the target-language response is clearly weaker or fails;
(ii) a false positive primarily caused by translation or formatting
artifacts; or (iii) ambiguous. Under this protocol, 85.2\% of examples
match the intended English-stronger/target-weaker pattern, 7.1\% are
translation/formatting false positives, and 7.7\% are ambiguous. These results support PPL-derived \textit{proxy-ci} as a practical
sample-selection signal despite its noise.

\section{Additional Method Details}
\label{app:method_details}

\subsection{Entropy-Normalized Jensen--Shannon Divergence}
\label{app:js}

For two routing distributions $\mathbf{p}$ and $\mathbf{q}$ over
$E$ experts, we first add a small constant to every component and
renormalize:
\begin{equation}
    \bar{p}_e
    =
    \frac{p_e+\epsilon}
         {\sum_{e'}(p_{e'}+\epsilon)},
    \qquad
    \bar{q}_e
    =
    \frac{q_e+\epsilon}
         {\sum_{e'}(q_{e'}+\epsilon)},
    \qquad
    \epsilon=10^{-12}.
\end{equation}
All logarithms used in this diagnostic are base two. Let
\begin{equation}
    \bar{\mathbf{m}}
    =
    \frac{\bar{\mathbf{p}}+\bar{\mathbf{q}}}{2}.
\end{equation}
The Jensen--Shannon divergence is computed as
\begin{equation}
    \mathrm{JS}_2
    (\bar{\mathbf{p}},\bar{\mathbf{q}})
    =
    \frac{1}{2}
    D_{\mathrm{KL},2}
    \left(
        \bar{\mathbf{p}}
        \,\middle\|\,
        \bar{\mathbf{m}}
    \right)
    +
    \frac{1}{2}
    D_{\mathrm{KL},2}
    \left(
        \bar{\mathbf{q}}
        \,\middle\|\,
        \bar{\mathbf{m}}
    \right),
\end{equation}
where $D_{\mathrm{KL},2}$ denotes KL divergence with base-2 logarithms.
We define the base-2 entropy as
\begin{equation}
    H_2(\mathbf{p})
    =
    -\sum_e p_e\log_2 p_e,
\end{equation}
and the mean entropy of the two routing distributions as
\begin{equation}
    \overline{H}_2
    (\bar{\mathbf{p}},\bar{\mathbf{q}})
    =
    \frac{1}{2}
    \left[
        H_2(\bar{\mathbf{p}})
        +
        H_2(\bar{\mathbf{q}})
    \right].
\end{equation}
The entropy-normalized divergence reported in the paper is
\begin{equation}
    D_{\mathrm{JS}}(\mathbf{p},\mathbf{q})
    =
    \frac{
        \mathrm{JS}_2
        (\bar{\mathbf{p}},\bar{\mathbf{q}})
    }{
        \log_2 E
        -
        \overline{H}_2
        (\bar{\mathbf{p}},\bar{\mathbf{q}})
    }.
\end{equation}
The denominator measures the gap between the maximum entropy attainable
over $E$ experts and the average entropy of the two routing
distributions. If this denominator is at most $10^{-10}$, the
implementation returns zero. No additional clipping is applied.

\subsection{Middle-Layer Identification}
\label{app:middle_layers}

The middle-layer interval is identified independently for each
model--task--target-language configuration. For a fixed configuration,
we first average the per-example routing divergence at every MoE layer:
\begin{equation}
    \mathrm{Div}^{(\ell)}
    =
    \frac{1}{N}
    \sum_{i=1}^{N}
    D_{i,\ell},
\end{equation}
where
\begin{equation}
    D_{i,\ell}
    =
    D_{\mathrm{JS}}
    \left(
        \mathbf{q}^{(\mathrm{en},\ell,i)},
        \mathbf{q}^{(\mathrm{tgt},\ell,i)}
    \right)
\end{equation}
is the entropy-normalized cross-lingual routing divergence for example
$i$ at layer $\ell$.

We then set the divergence threshold to the median of the layer-wise
means:
\begin{equation}
    \tau
    =
    \operatorname{median}_{\ell}
    \left\{
        \mathrm{Div}^{(\ell)}
    \right\}.
\end{equation}
Layers satisfying
\begin{equation}
    \mathrm{Div}^{(\ell)}
    \leq
    \tau
\end{equation}
are retained, and the selected middle-layer range
$\mathcal{L}_{\mathrm{mid}}$ is defined as the longest contiguous
interval among these layers. If multiple intervals have the same maximum
length, the earliest interval in layer order is retained. Consequently,
$\mathcal{L}_{\mathrm{mid}}$ is determined separately for every
model--task--target-language configuration and is not shared across
target languages.

Table~\ref{tab:middle_ranges} reports the selected ranges used in the
core experiments. All ranges are inclusive, zero-based absolute
transformer-layer indices.

\begin{table}[t]
\caption{Selected middle-layer ranges used in the core experiments.
All ranges are inclusive, zero-based absolute transformer-layer indices.}
\label{tab:middle_ranges}
\centering
\small
\renewcommand{\arraystretch}{1.05}
\setlength{\tabcolsep}{2.5pt}

\begin{tabular}{llcccccc}
\toprule
\textbf{Model} & \textbf{Task}
& \textbf{ar} & \textbf{bn} & \textbf{zh}
& \textbf{fr} & \textbf{ja} & \textbf{es} \\
\midrule

\multirow{3}{*}{\tolmoe}
& GSM8K
& 5--11 & 4--9 & 5--11 & 4--11 & 4--10 & 4--11 \\
& LogicBench
& 4--11 & 4--11 & 4--11
& 4--11 & 4--11 & 4--11 \\
& MMLU
& 6--12 & 6--13 & 6--13
& 5--12 & 4--11 & 4--11 \\

\midrule

\multirow{3}{*}{\tqwen}
& GSM8K
& 5--14 & 5--15 & 5--13 & 5--15 & 5--13 & 4--13 \\
& LogicBench
& 5--11 & 5--11 & 4--12
& 4--11 & 4--15 & 4--12 \\
& MMLU
& 8--16 & 8--16 & 8--16
& 6--16 & 8--16 & 7--16 \\

\midrule

\multirow{3}{*}{\tdeepseek}
& GSM8K
& 6--16 & 6--14 & 7--15
& 6--15 & 6--16 & 7--14  \\
& LogicBench
& 9--16 & 11--16 & 9--16
& 10--15 & 9--16 & 8--15 \\
& MMLU
& 10--17 & 9--16 & 9--15
& 11--16 & 9--15 & 10--16 \\

\bottomrule
\end{tabular}
\end{table}

\subsection{Derivation of the Routing Alignment Decomposition}
\label{app:alignment_decomposition}

We provide the derivation of the routing-alignment decomposition used in
Section~\ref{sec:method}. For clarity, we omit the sample and layer
indices $(i,\ell)$ throughout this section.

Let $\mathcal{E}^{\mathrm{task}}$ denote the task-expert set and define
the total routing mass assigned to these experts as
\begin{equation}
    m^{(\text{lang})}
    =
    \sum_{e\in\mathcal{E}^{\mathrm{task}}}
    q_e^{(\text{lang})}.
\end{equation}
The normalized routing distribution within the task-expert set is
\begin{equation}
    \tilde{q}_e^{(\text{lang})}
    =
    \frac{
        q_e^{(\text{lang})}
    }{
        m^{(\text{lang})}
    },
    \qquad
    e\in\mathcal{E}^{\mathrm{task}},
\end{equation}
such that
\begin{equation}
    q_e^{(\text{lang})}
    =
    m^{(\text{lang})}
    \tilde{q}_e^{(\text{lang})},
    \qquad
    \sum_{e\in\mathcal{E}^{\mathrm{task}}}
    \tilde{q}_e^{(\text{lang})}
    =
    1.
\end{equation}

Recall that RA-MoE constructs the collapsed routing distribution
\begin{equation}
    \mathbf{r}^{(\text{lang})}
    =
    \left[
        \left(
            q_e^{(\text{lang})}
        \right)_{e\in\mathcal{E}^{\mathrm{task}}},
        \;
        1-m^{(\text{lang})}
    \right],
\end{equation}
which preserves the routing probability assigned to each identified task
expert while aggregating all remaining experts into a single residual
component.

The KL divergence between the English and target-language collapsed
distributions can be expanded as
\begin{align}
&D_{\mathrm{KL}}
\left(
    \mathbf{r}^{(\mathrm{en})}
    \,\Big\|\,
    \mathbf{r}^{(\mathrm{tgt})}
\right)
\nonumber\\
={}&
\sum_{e\in\mathcal{E}^{\mathrm{task}}}
q_e^{(\mathrm{en})}
\log
\frac{
    q_e^{(\mathrm{en})}
}{
    q_e^{(\mathrm{tgt})}
}
+
\left(
    1-m^{(\mathrm{en})}
\right)
\log
\frac{
    1-m^{(\mathrm{en})}
}{
    1-m^{(\mathrm{tgt})}
}.
\label{eq:app_kl_expand}
\end{align}

Substituting
\begin{equation}
    q_e^{(\text{lang})}
    =
    m^{(\text{lang})}
    \tilde{q}_e^{(\text{lang})}
\end{equation}
into the first term gives
\begin{align}
&\sum_{e\in\mathcal{E}^{\mathrm{task}}}
m^{(\mathrm{en})}
\tilde{q}_e^{(\mathrm{en})}
\log
\frac{
    m^{(\mathrm{en})}
    \tilde{q}_e^{(\mathrm{en})}
}{
    m^{(\mathrm{tgt})}
    \tilde{q}_e^{(\mathrm{tgt})}
}
\nonumber\\
={}&
m^{(\mathrm{en})}
\sum_{e\in\mathcal{E}^{\mathrm{task}}}
\tilde{q}_e^{(\mathrm{en})}
\left[
    \log
    \frac{
        m^{(\mathrm{en})}
    }{
        m^{(\mathrm{tgt})}
    }
    +
    \log
    \frac{
        \tilde{q}_e^{(\mathrm{en})}
    }{
        \tilde{q}_e^{(\mathrm{tgt})}
    }
\right]
\nonumber\\
={}&
m^{(\mathrm{en})}
\log
\frac{
    m^{(\mathrm{en})}
}{
    m^{(\mathrm{tgt})}
}
+
m^{(\mathrm{en})}
D_{\mathrm{KL}}
\left(
    \tilde{\mathbf{q}}^{(\mathrm{en})}
    \,\Big\|\,
    \tilde{\mathbf{q}}^{(\mathrm{tgt})}
\right),
\label{eq:app_task_kl}
\end{align}
where the final equality follows from
\begin{equation}
    \sum_{e\in\mathcal{E}^{\mathrm{task}}}
    \tilde{q}_e^{(\mathrm{en})}
    =
    1.
\end{equation}

We next define the binary distribution describing routing mass inside
and outside the task-expert set:
\begin{equation}
    \mathbf{b}^{(\text{lang})}
    =
    \left[
        m^{(\text{lang})},
        1-m^{(\text{lang})}
    \right].
\end{equation}
Its KL divergence is
\begin{align}
D_{\mathrm{KL}}
\left(
    \mathbf{b}^{(\mathrm{en})}
    \,\Big\|\,
    \mathbf{b}^{(\mathrm{tgt})}
\right)
={}&
m^{(\mathrm{en})}
\log
\frac{
    m^{(\mathrm{en})}
}{
    m^{(\mathrm{tgt})}
}
\nonumber\\
&+
\left(
    1-m^{(\mathrm{en})}
\right)
\log
\frac{
    1-m^{(\mathrm{en})}
}{
    1-m^{(\mathrm{tgt})}
}.
\label{eq:app_mass_kl}
\end{align}

Combining Equations~\ref{eq:app_kl_expand},
\ref{eq:app_task_kl}, and \ref{eq:app_mass_kl} yields
\begin{equation}
D_{\mathrm{KL}}
\left(
    \mathbf{r}^{(\mathrm{en})}
    \,\Big\|\,
    \mathbf{r}^{(\mathrm{tgt})}
\right)
=
D_{\mathrm{KL}}
\left(
    \mathbf{b}^{(\mathrm{en})}
    \,\Big\|\,
    \mathbf{b}^{(\mathrm{tgt})}
\right)
+
m^{(\mathrm{en})}
D_{\mathrm{KL}}
\left(
    \tilde{\mathbf{q}}^{(\mathrm{en})}
    \,\Big\|\,
    \tilde{\mathbf{q}}^{(\mathrm{tgt})}
\right).
\end{equation}

The first term measures the discrepancy in the total routing mass
assigned to the task-expert set, whereas the second measures the
discrepancy in the normalized allocation within that set. The factor
$m^{(\mathrm{en})}$ follows directly from the KL decomposition and
weights within-set allocation differences more strongly when the
successful English routing assigns greater probability mass to the
identified task experts.

\section{Experimental Implementation}
\label{app:impl}

\subsection{Model Details}
\label{app:models}

We evaluate RA-MoE on three pretrained MoE models spanning different
architectures and parameter scales. Table~\ref{tab:model_arch} summarizes
their key architectural properties.

\begin{table}[t]
\caption{Architectural summary of the three MoE models. $^\dagger$Qwen1.5-MoE and DS-V2-Lite include four and two shared experts activated for every token in addition to the routed experts, respectively. DS-V2-Lite has one dense layer followed by 26 MoE layers.}
\label{tab:model_arch}
\centering
\small
\renewcommand{\arraystretch}{1.12}
\setlength{\tabcolsep}{4.5pt}

\begin{tabular}{lcccccc}
\toprule
\textbf{Model} & \textbf{Total params} & \textbf{Active params}
& \textbf{Layers} & \textbf{MoE layers} & \textbf{Routed experts} & \textbf{Top-$k$} \\
\midrule
OLMoE-1B-7B        & 7B   & 1B   & 16 & 16 & 64 & 8 \\
Qwen1.5-MoE        & 14.3B & 2.7B & 24 & 24 & 60+4$^\dagger$ & 4+4$^\dagger$ \\
DS-V2-Lite         & 16B  & 2.4B & 27 & 26 & 64+2$^\dagger$ & 6+2$^\dagger$ \\
\bottomrule
\end{tabular}
\end{table}

We use the following model variants in all main experiments:
\texttt{allenai/OLMoE-1B-7B-0924-SFT},
\texttt{Qwen/Qwen1.5-MoE-A2.7B}, and
\texttt{deepseek-ai/DeepSeek-V2-Lite-Chat}.

\subsection{Training Hyperparameters}
\label{app:hyperparams}

All training-based methods use LoRA~\citep{hu2022lora} on the up and down
projections of each FFN/MoE expert layer, with the remaining model
parameters frozen. SFT, RISE, and RA-MoE use the same optimization
schedule and data order within each comparison.

\begin{table}[t]
\caption{Hyperparameter settings shared across the core experiments.}
\label{tab:hyperparams}
\centering
\small
\renewcommand{\arraystretch}{1.12}
\setlength{\tabcolsep}{6pt}
\begin{tabular}{lc}
\toprule
\textbf{Hyperparameter} & \textbf{Value} \\
\midrule
\multicolumn{2}{l}{\textit{LoRA settings}} \\
LoRA rank $r$            & 16 \\
LoRA $\alpha$            & 32 \\
LoRA dropout              & 0.05 \\
Target modules            & up/down projections \\
\midrule
\multicolumn{2}{l}{\textit{Optimization}} \\
Optimizer                 & AdamW \\
Learning rate             & $2\times10^{-5}$ \\
LR scheduler              & Linear \\
Warmup ratio              & 0.03 \\
Weight decay              & 0.0 \\
Gradient clipping         & None \\
Epochs                     & 1 \\
Precision                  & bfloat16 \\
\midrule
\multicolumn{2}{l}{\textit{RA-MoE specific}} \\
Alignment weight $\lambda$ & 1.0 \\
Task experts per layer $K$  & 8 \\
Middle-layer threshold $\tau$ & median (50th percentile) \\
General corpus (Stage~2)    & FLORES-200 \\
Routing-profiling batch size & 32 \\
\bottomrule
\end{tabular}
\end{table}

Table~\ref{tab:hyperparams_per_task} records settings that depend on the
model/task pair.

\begin{table}[t]
\caption{Per-model, per-task batch-size and sequence-length settings for
the core experiments. Effective batch size includes gradient accumulation
across four GPUs.}
\label{tab:hyperparams_per_task}
\centering
\small
\renewcommand{\arraystretch}{1.08}
\setlength{\tabcolsep}{5pt}

\begin{tabular}{llccc}
\toprule
& & \textbf{GSM8K} & \textbf{LogicBench} & \textbf{MMLU} \\
\midrule
\multirow{4}{*}{\olmoe}
& Micro-batch size       & 16  & 8    & 8 \\
& Gradient accum. steps  & 4   & 8    & 8 \\
& Effective batch size   & 256 & 256  & 256 \\
& Max sequence length    & 512 & 2048 & 2048 \\
\midrule
\multirow{4}{*}{\tdeepseek}
& Micro-batch size       & 4   & 3    & 3 \\
& Gradient accum. steps  & 4   & 2    & 2 \\
& Effective batch size   & 64  & 24   & 24 \\
& Max sequence length    & 512 & 2048 & 2048 \\
\midrule
\multirow{4}{*}{\tqwen}
& Micro-batch size       & 4   & 3    & 3 \\
& Gradient accum. steps  & 4   & 2    & 2 \\
& Effective batch size   & 64  & 24   & 24 \\
& Max sequence length    & 512 & 2048 & 2048 \\
\bottomrule
\end{tabular}
\end{table}

\subsection{Routing-Alignment Implementation}
\label{app:routing_impl}

For a minibatch $\mathcal{B}$, let $\mathcal{B}_{\mathrm{ci}}\subseteq\mathcal{B}$
denote the examples selected for routing alignment. The implementation
averages the KL divergence over the selected examples and middle layers:
\begin{equation}
\widehat{\mathcal{L}}_{\mathrm{align}}
=
\frac{1}{|\mathcal{B}_{\mathrm{ci}}|}
\sum_{i\in\mathcal{B}_{\mathrm{ci}}}
\frac{1}{|\mathcal{L}_{\mathrm{mid}}|}
\sum_{\ell\in\mathcal{L}_{\mathrm{mid}}}
D_{\mathrm{KL}}\!\left(
\mathbf{r}^{(\mathrm{en},\ell,i)}
\,\Big\|\,
\mathbf{r}^{(\mathrm{tgt},\ell,i)}
\right).
\end{equation}
When a local minibatch contains no selected examples, we set
$\widehat{\mathcal{L}}_{\mathrm{align}}=0$. The English routing
reference is fixed, while the target-language routing distribution is
computed from the current SFT forward pass. The final objective is
$\mathcal{L}_{\mathrm{CE}}+\lambda\widehat{\mathcal{L}}_{\mathrm{align}}$;
all examples contribute to $\mathcal{L}_{\mathrm{CE}}$, whereas only
selected \textit{ci} (or \textit{proxy-ci} in the IFEval extension)
examples contribute to the alignment term.

\noindent\textbf{Order of aggregation.}
For each sample and selected middle layer, we first average router
probabilities over assistant-response tokens to obtain a sequence-level
routing distribution. We then construct the collapsed routing
distribution $\mathbf{r}$ and compute the KL divergence between the
English and target-language distributions. The resulting KL values are
averaged across selected middle layers and \textit{ci} examples in the
minibatch.

\noindent\textbf{Distributed training.}
For each local micro-batch, each worker identifies its local selected
examples and averages the per-example alignment losses over this subset.
If no selected examples are present, the local alignment loss is set to
zero. Standard distributed data parallelism then averages gradients
equally across workers, without reweighting by the number of selected
examples on each worker. Consequently, the resulting routing gradient
corresponds to an average of worker-local means rather than an exact mean
over all selected examples in the global batch. With gradient
accumulation, this computation is performed independently for each
micro-batch.

\subsection{Computational Infrastructure}
\label{app:hardware}

All experiments are conducted on a server equipped with 8$\times$
RTX~5880 Ada (48\,GB each), with four GPUs used for each training run. We use DeepSpeed ZeRO-2 for distributed training,
bfloat16 mixed precision, and Flash Attention~2 for memory-efficient
attention computation.

\subsection{Evaluation Settings}
\label{app:evaluation}

The benchmark splits, metrics, and few-shot settings are summarized in
Table~\ref{tab:datasets}. All evaluations use vLLM. GSM8K and LogicBench
use greedy decoding, with answers scored using the task-specific
procedures described in Section~\ref{app:correctness}. MMLU is evaluated
0-shot using constrained multiple-choice scoring over
$\{\texttt{A},\texttt{B},\texttt{C},\texttt{D}\}$, selecting the option
with the highest next-token log probability.

For each model--method setting, we repeat evaluation five times using
random seeds $\{42,123,456,789,2026\}$ and report the mean accuracy
and standard deviation across these runs.

\subsection{Baseline Reimplementation Details}
\label{app:baselines}

\noindent\textbf{Routing Steering (RS).}
Routing Steering~\citep{bandarkar2026multilingualroutingmixtureofexperts}
is applied at inference time without updating model parameters. We follow
the original implementation and default settings, using the same
middle-layer region as RA-MoE and without task- or language-specific
hyperparameter tuning.

\noindent\textbf{RISE.}
RISE~\citep{zheng2026unveilinglanguageroutingisolation} is reproduced
following the original implementation and default expert-selection
procedure. The selected language-specific expert subnetwork is adapted
using the same training data, optimizer, learning rate, and training
schedule as RA-MoE, without task- or language-specific hyperparameter
tuning.

\section{Additional Analysis Details}
\label{app:analysis_details}

\subsection{Cross-Language Transfer Experiments}
\label{app:transfer_details}

\noindent\textbf{Reference-language transfer.}
Section~\ref{sec:cross_lingual_transfer} evaluates whether successful
routing supervision must come from English. Both experiments target
Japanese on GSM8K. For DeepSeek-V2-Lite-Chat, the reference language is
English, Chinese, or Spanish; for Qwen1.5-MoE, the reference is English,
French, or Spanish. Each non-English reference is first adapted with
standard SFT to establish sufficient task competence. We then identify
its task experts analogously to the English-reference setting and use
successful routing behavior in that language as the alignment reference.
For each reference--target pair, the correctness groups are recomputed
independently, so \textit{ci} denotes examples on which the current
reference language is correct and Japanese is incorrect.

\noindent\textbf{Target-side task-expert transfer.}
For RA-MoE$^\dagger$, we identify the middle-layer range, per-layer
task-expert identities, and per-example English reference routing using
the Chinese target-language configuration for each model--task pair.
These components are then reused unchanged when fine-tuning Arabic or
Bengali, while all other training settings remain the same. Thus,
RA-MoE$^\dagger$ directly tests whether a task-routing configuration
identified for one target language can be transferred to another without
re-running the identification or reference-routing stages.

\subsection{Mechanistic Analysis of Language-Shared Task Routing}
\label{app:mechanistic_details}

We compare GSM8K task routing across English, Chinese, and Spanish on
DeepSeek-V2-Lite-Chat. For routing-pattern comparisons, we use a shared
set of 12{,}388 parallel examples that are correctly solved in all three
reference languages but incorrectly solved in Japanese. Specifically,
this set is the intersection of the pair-specific English--Japanese,
Chinese--Japanese, and Spanish--Japanese \textit{ci} subsets. This
controls sample composition across reference languages.

\noindent\textbf{Task-expert identity.}
For task-expert sets $\mathcal{E}^{(\ell)}_A$ and
$\mathcal{E}^{(\ell)}_B$, we compute layer-wise Jaccard similarity
\begin{equation}
J^{(\ell)}(A,B)
=
\frac{|\mathcal{E}^{(\ell)}_A\cap\mathcal{E}^{(\ell)}_B|}
     {|\mathcal{E}^{(\ell)}_A\cup\mathcal{E}^{(\ell)}_B|},
\end{equation}
and average over the selected middle layers. The observed pairwise
cross-language similarities are 0.4182 (en--zh), 0.4096 (en--es), and
0.5467 (zh--es), yielding a mean of approximately 0.458. As a cross-task
control, English GSM8K vs. English MMLU yields a mean Jaccard similarity
of 0.0543.

\noindent\textbf{Random-expert control.}
For $E=64$ experts and $K=8$ selected experts per layer, we draw
$10^6$ pairs of random size-$K$ subsets and compute their Jaccard
similarity. The random baseline has mean Jaccard 0.0706. The empirical
probability of obtaining a random Jaccard at least as large as the
smallest observed cross-language value (0.4096) is
$3.57\times10^{-4}$.

\noindent\textbf{Task-expert usage.}
We additionally compare the total routing mass assigned to task experts
and the normalized routing distribution within the shared task-expert
set. The mean task-expert masses are 0.1837 (en), 0.1826 (zh), and
0.1752 (es). Pairwise mean absolute mass differences are 0.00923
(en--zh), 0.00844 (en--es), and 0.01022 (zh--es). The corresponding
within-set JS divergences are 0.00281, 0.00226, and 0.00333.

\begin{table}[t]
\caption{Pairwise routing consistency across successful reference languages on DeepSeek-V2-Lite-Chat (GSM8K).}
\label{tab:mechanistic_pairwise}
\centering
\small
\renewcommand{\arraystretch}{1.08}
\setlength{\tabcolsep}{4.5pt}
\begin{tabular}{lccc}
\toprule
\textbf{Pair} & \textbf{Jaccard} & \textbf{Mass diff.} & \textbf{Within-set JS} \\
\midrule
en--zh & 0.4182 & 0.00923 & 0.00281 \\
en--es & 0.4096 & 0.00844 & 0.00226 \\
zh--es & 0.5467 & 0.01022 & 0.00333 \\
\bottomrule
\end{tabular}
\end{table}

\section{Computational Cost Analysis}
\label{app:compute_analysis}

We analyze RA-MoE overhead in terms of preparatory inference time,
training-time memory, and additional routing-alignment arithmetic. Stage~1
uses autoregressive inference to categorize training examples and Stage~2
uses teacher-forced forward passes to profile routing. Neither stage
requires gradient computation. Stage~3 uses the same model forward pass
as standard SFT and adds the routing-alignment loss to already available
router outputs; it does not require an additional model forward/backward
pass.

\subsection{Training Time Analysis}
\label{app:training_time}

RA-MoE introduces two preparatory stages before fine-tuning: Stage~1
constructs the correctness-based sample groups, and Stage~2 profiles
routing behavior to identify the selected middle layers and task experts
and to obtain the reference routing signals. Both stages require only
inference, without gradient computation. During Stage~3, the routing
alignment loss is computed from router probabilities already produced by
the SFT forward pass and therefore does not require an additional model
forward or backward pass.

Table~\ref{tab:training_time} reports wall-clock time for all three
models and core tasks. Stage~3 training time remains comparable to
standard SFT across all settings. When the preparatory stages are run
separately for each target-language configuration, the Stage~1+2 cost is
at most 16.7\% of Stage~3. Reusing the middle-layer range, per-layer
task-expert identities, and English reference routing across target
languages further reduces this overhead to at most 10.0\%.

\begin{table}[t]
\caption{Wall-clock time (hours) decomposed by stage. S1+2 denotes the
two preparatory stages and S3 denotes fine-tuning. \textit{Per-lang.}
runs the routing-identification pipeline separately for each
target-language configuration, whereas \textit{reuse} transfers the
identified middle-layer range, task experts, and English reference
routing across target languages.}
\label{tab:training_time}
\centering
\small
\renewcommand{\arraystretch}{1.05}
\setlength{\tabcolsep}{2.5pt}

\begin{tabular}{llrrrrrr}
\toprule
& & \multicolumn{2}{c}{\textbf{GSM8K}}
& \multicolumn{2}{c}{\textbf{LogicBench}}
& \multicolumn{2}{c}{\textbf{MMLU}} \\
\cmidrule(lr){3-4}
\cmidrule(lr){5-6}
\cmidrule(lr){7-8}
\textbf{Model} & \textbf{Method}
& \textbf{S1+2} & \textbf{S3}
& \textbf{S1+2} & \textbf{S3}
& \textbf{S1+2} & \textbf{S3} \\
\midrule

\multirow{3}{*}{\tolmoe}
& SFT
& --  & 3.0
& --  & 1.2
& --  & 3.5 \\
& RA-MoE (per-lang.)
& 0.5 & 3.0
& 0.2 & 1.2
& 0.5 & 3.5 \\
& RA-MoE (reuse)
& 0.3 & 3.0
& 0.1 & 1.2
& 0.3 & 3.5 \\

\midrule

\multirow{3}{*}{\tqwen}
& SFT
& --  & 15.5
& --  & 1.5
& --  & 18.5 \\
& RA-MoE (per-lang.)
& 2.0 & 15.0
& 0.2 & 1.5
& 3.0 & 18.5 \\
& RA-MoE (reuse)
& 1.0 & 15.0
& 0.1 & 1.5
& 1.5 & 18.5 \\

\midrule

\multirow{3}{*}{\tdeepseek}
& SFT
& --  & 12.0
& --  & 1.5
& --  & 18.0 \\
& RA-MoE (per-lang.)
& 1.5 & 11.5
& 0.2 & 1.5
& 3.0 & 18.0 \\
& RA-MoE (reuse)
& 0.8 & 11.5
& 0.1 & 1.5
& 1.5 & 18.0 \\

\bottomrule
\end{tabular}
\end{table}

\subsection{Memory Overhead}
\label{app:memory_analysis}

RA-MoE introduces no additional trainable model parameters beyond the
LoRA adapters used by standard SFT. Under the matched OLMoE
configuration, both SFT and RA-MoE contain 100{,}663{,}296 trainable
parameters. The stored English reference routing distributions are kept
in CPU memory as float32 tensors and transferred to the GPU only for the
selected examples in the current batch.

During Stage~3, RA-MoE additionally retains router outputs from the
selected middle layers to compute the alignment loss. In the available
training logs, SFT and RA-MoE show the same allocated GPU memory after
optimizer initialization (13.26~GB in the matched OLMoE configuration).
Thus, RA-MoE does not introduce additional model, LoRA, or optimizer
states; its extra runtime memory is limited to the routing references and
intermediate router outputs required by the alignment objective.

\subsection{Alignment-Loss Computation}
\label{app:flops_analysis}

Once the router probabilities are available from the standard forward
pass, RA-MoE computes the alignment loss only for \textit{ci} examples
and selected middle layers. Since routing probabilities are first
averaged over response tokens, the KL divergence is computed once per
selected example and layer rather than once per token. The collapsed
distribution contains only $K+1$ entries---the $K$ task experts and one
residual component---so, following the same coarse arithmetic accounting
used for the base-model computation, the alignment cost scales as
\begin{equation}
    \mathrm{FLOPs}_{\mathrm{align}}
    \approx
    2
    |\mathcal{B}_{\mathrm{ci}}|
    |\mathcal{L}_{\mathrm{mid}}|
    (K+1).
\end{equation}

To provide a conservative numerical comparison, we use OLMoE with the
same representative configuration as the base-computation estimate:
$B=4096$ tokens, $L=16$ MoE layers, $k_{\mathrm{act}}=8$ activated
experts per token, $d_m=2048$, $d_e=1024$, and LoRA rank $r=16$.
The expert FFN and LoRA computations are
\begin{align}
    \mathrm{FLOPs}_{\mathrm{base}}
    &=
    6BLk_{\mathrm{act}}d_md_e
    =
    6597.1~\mathrm{GFLOPs},\\
    \mathrm{FLOPs}_{\mathrm{LoRA}}
    &=
    4BLk_{\mathrm{act}}r(d_m+d_e)
    =
    103.1~\mathrm{GFLOPs}.
\end{align}
For RA-MoE, $K=8$. Even under the deliberately conservative assumption
that all $B=4096$ token positions correspond to distinct selected
examples and that alignment is applied to all $L=16$ MoE layers, the
alignment computation is bounded by
\begin{equation}
    \mathrm{FLOPs}_{\mathrm{align}}
    \leq
    2BL(K+1)
    \approx
    0.00118~\mathrm{GFLOPs}.
\end{equation}
This upper bound is only about $1.8\times10^{-5}\%$ of the approximately
$6700.1$ GFLOPs required by the expert and LoRA computations. The actual
overhead is smaller because alignment is applied only to \textit{ci}
examples and selected middle layers. Thus, the arithmetic cost of the routing-alignment objective is
negligible relative to the expert and LoRA computations of the
underlying MoE model.

\end{document}